\documentclass[9.5pt,journal,cspaper,compsoc]{IEEEtran}
\ifCLASSOPTIONcompsoc
  \usepackage[nocompress]{cite}
\else
  \usepackage{cite}
\fi
\usepackage{graphicx}
\usepackage{amsmath}
\usepackage{amssymb}
\usepackage{booktabs}
\usepackage{epsfig}
\usepackage{graphicx}
\usepackage{multirow}
\usepackage{makecell}
\usepackage{url}
\usepackage[table]{xcolor}
\usepackage{caption}
\usepackage{subcaption}
\usepackage{enumerate}
\usepackage{inconsolata}
\usepackage{diagbox}
\usepackage{overpic}

\definecolor{mygray}{gray}{0.87}
\newcommand\ChangeRT[1]{\noalign{\hrule height #1}}
\usepackage{float}

\ifCLASSINFOpdf

\else

\fi

\begin{document}

\title{Prophet: Prompting Large Language Models with Complementary Answer Heuristics for Knowledge-based Visual Question Answering}

\author{Zhou~Yu~\IEEEmembership{Member,~IEEE},~
        Xuecheng~Ouyang,~
        Zhenwei~Shao,~\\
        Meng~Wang~\IEEEmembership{Fellow,~IEEE},~
        Jun Yu~\IEEEmembership{Senior Member,~IEEE}
\IEEEcompsocitemizethanks{
\IEEEcompsocthanksitem This work was supported in part by the National Natural Science Foundation of China under Grants 62422204, 62125201, U24B20174, 62072147, 62020106007, and in part by the Zhejiang Provincial Natural Science Foundation of China under Grant LR22F020001, LDT23F02025F02. (Corresponding author: Jun Yu.)
\IEEEcompsocthanksitem Z. Yu, Z. Shao are with the Key Laboratory of Complex Systems Modeling and Simulation, the School of Computer Science, Hangzhou Dianzi University, China. (e-mail: yuz@hdu.edu.cn; shaozw@hdu.edu.cn)
\IEEEcompsocthanksitem X. Ouyang is with the HDU-ITMO Joint Institute, Hangzhou Dianzi University, China. (e-mail: ouyangxc@hdu.edu.cn)
\IEEEcompsocthanksitem M. Wang is with the School of Computer Science and Information Engineering, Hefei University of Technology, China. (e-mail: eric.mengwang@gmail.com)
\IEEEcompsocthanksitem J. Yu is with the School of Intelligence Science and Engineering, Harbin
Institute of Technology, Shenzhen, China (e-mail: yujun@hit.edu.cn)
}
}

\markboth{IEEE Transactions on Pattern Analysis and Machine Intelligence}%
{Yu \MakeLowercase{\textit{et al.}}: Prophet: Prompting Large Language Models with Complementary Answer Heuristics for KBVQA}

\IEEEtitleabstractindextext{
\begin{abstract}
	Knowledge-based visual question answering (VQA) requires external knowledge beyond the image to answer the question. Early studies retrieve required knowledge from explicit knowledge bases (KBs), which often introduces irrelevant information to the question, hence restricting the performance of their models. Recent works have resorted to using a powerful large language model (LLM) as an implicit knowledge engine to acquire the necessary knowledge for answering. Despite the encouraging results achieved by these methods, we argue that they have not fully activated the capacity of the \emph{blind} LLM as the provided textual input is insufficient to depict the required visual information to answer the question. In this paper, we present Prophet---a conceptually simple, flexible, and general framework designed to \textbf{\underline{pro}}m\textbf{\underline{p}}t LLM with answer \textbf{\underline{he}}uris\textbf{\underline{t}}ics for knowledge-based VQA. Specifically, we first train a vanilla VQA model on a specific knowledge-based VQA dataset without external knowledge. After that, we extract two types of complementary answer heuristics from the VQA model: answer candidates and answer-aware examples. The two types of answer heuristics are jointly encoded into a formatted prompt to facilitate the LLM's understanding of both the image and question, thus generating a more accurate answer. By incorporating the state-of-the-art LLM GPT-3 \cite{gpt3}, Prophet significantly outperforms existing state-of-the-art methods on four challenging knowledge-based VQA datasets. Prophet is general that can be instantiated with the combinations of different VQA models (\emph{i.e.}, both discriminative and generative ones) and different LLMs (\emph{i.e.}, both commercial and open-source ones). Moreover, Prophet can also be integrated with modern large multimodal models in different stages, which is named Prophet++, to further improve the capabilities on knowledge-based VQA tasks.
	
\end{abstract}

}

\maketitle

\IEEEdisplaynontitleabstractindextext

\IEEEpeerreviewmaketitle

\IEEEraisesectionheading{\section{Introduction}\label{sec:introduction}}
\IEEEPARstart{R}ecent advances in multimodal learning have achieved remarkable progress in various vision-language tasks, including visual captioning \cite{butd,mvmt}, visual grounding \cite{hong2019learning, deng2022visual, deng2023transvg}, and visual question answering (VQA) \cite{ban,mcan}. Among these tasks, VQA poses unique challenges by requiring machines to answer free-form questions through reasoning about given images. 
Benefiting from large-scale vision-language pretraining \cite{lxmert,li2020oscar,deng2023transvg}, the state-of-the-art methods have even surpassed human level on several representative benchmarks \cite{flamingo,ofa}. Despite the success of these methods, their reasoning abilities are far from satisfactory, especially when \emph{external knowledge} is required to answer the questions. In this situation, the task of knowledge-based VQA is introduced to validate models' abilities to leverage external knowledge. Early knowledge-based VQA benchmarks additionally provide structured knowledge bases (KBs) and annotate required knowledge facts for all the questions \cite{fvqa,kbvqa}. More recently, benchmarks emphasizing on \emph{open-domain} knowledge have been established \cite{okvqa,a-okvqa}, which means KBs are no longer provided and any external knowledge resource can be used for answering. We focus on the task with open-domain knowledge in this paper.

A straightforward solution for knowledge-based VQA is to {retrieve} knowledge entries from explicit KBs, \emph{e.g.}, Wikipedia and ConceptNet \cite{conceptnet}. Then, a KB-augmented VQA model performs joint reasoning over the retrieved knowledge, image, and question to predict the answer \cite{krisp,mavex,trig}. However, the performance of these retrieval-based approaches is limited for two reasons: (i) the required knowledge may not be successfully retrieved from the KBs; and (ii) even if the required knowledge is retrieved, plenty of irrelevant knowledge is inevitably introduced, which hampers the learning of VQA models.

\captionsetup[subfigure]{font=small}
\begin{figure}
	\begin{center}
		\begin{overpic}[width=\linewidth]{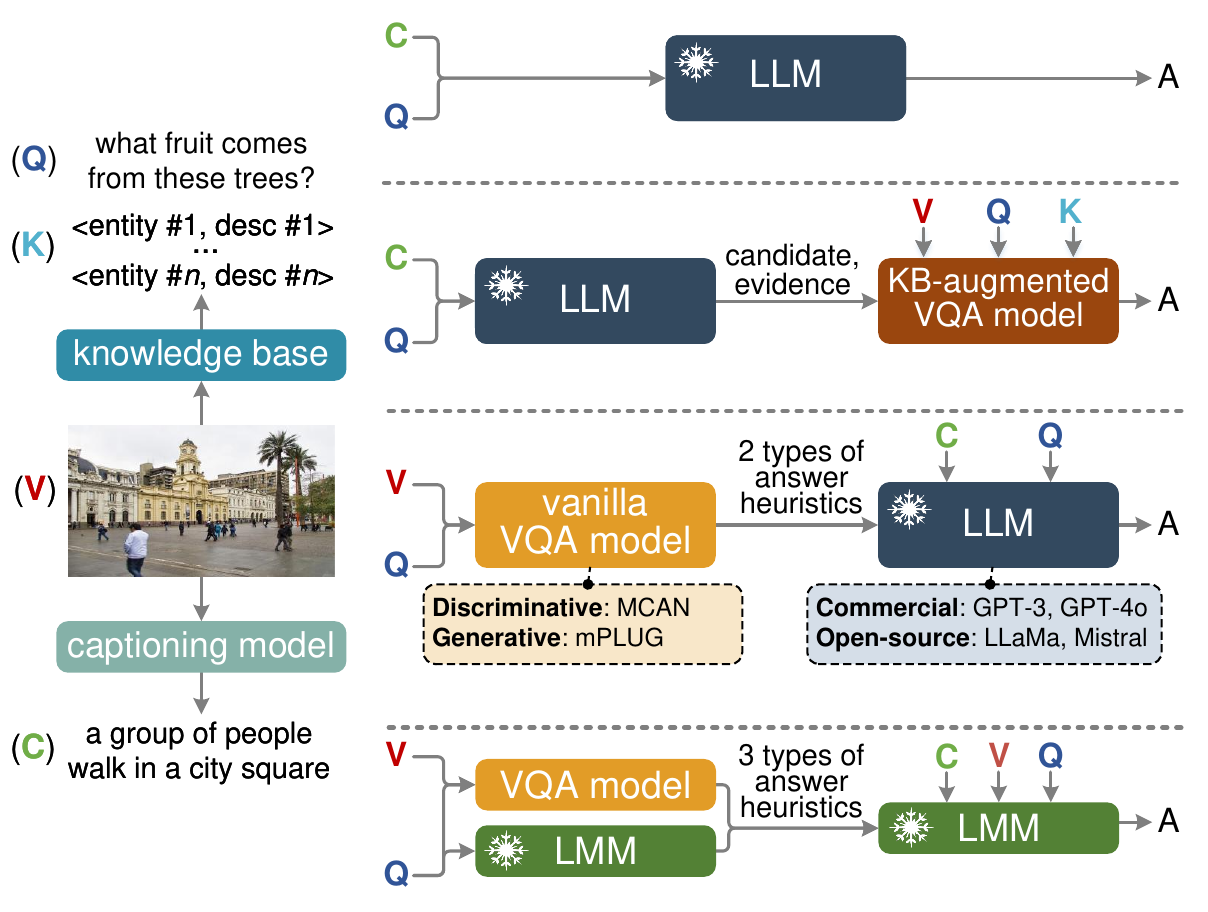}
			\put(57, 61.5){ \footnotesize {PICa \cite{pica}}}
			\put(47, 43){ \footnotesize{KAT \cite{kat} / REVIVE \cite{revive}} }
			\put(58, 16.5){ \small {\textbf{Prophet}}}
			\put(57, -0.5){ \small {\textbf{Prophet++}}}
		\end{overpic}
		\caption{Conceptual comparisons of three knowledge-based VQA frameworks using a frozen LLM model, \emph{e.g.}, GPT-3 \cite{gpt3}. While PICa \cite{pica}, KAT \cite{kat}, and REVIVE \cite{revive} directly feed the caption (C) and question (Q) into the LLM as the prompt, we argue that the information they provide for the LLM is insufficient thus cannot fully activate the LLM's potential. In contrast, our Prophet learns a vanilla VQA model without external knowledge to produce \emph{answer heuristics}, which endows the LLM with richer and more task-specific information for answer prediction. 
		In contrast to the counterparts that resort to specific VQA models and LLMs, Prophet is general that can be instantiated with the combinations of different VQA models (\emph{i.e.}, both discriminative and generative ones) and different LLMs (\emph{i.e.}, both commercial and open-source ones). Moreover, Prophet can also be integrated with large multimodal models (LMMs) in different stages, which is termed Prophet++, to further improve the capabilities on knowledge-based VQA tasks.}
		\label{fig:intro}
	\end{center}
	\vspace{-15pt}
\end{figure}
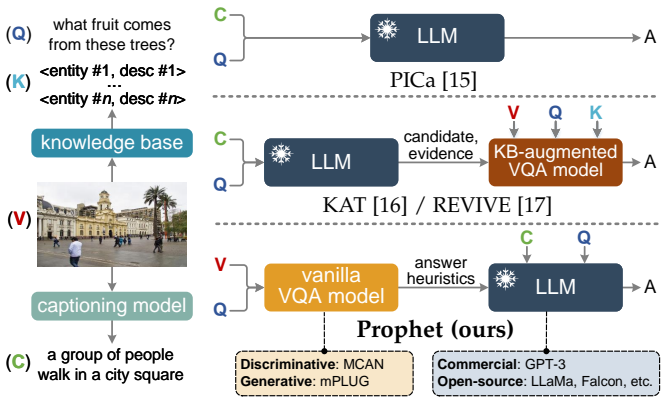

Apart from those studies using explicit KBs, another line of research resorts to pretrained large language models (LLMs), \emph{e.g.}, GPT-3 \cite{gpt3}, as implicit knowledge engines for knowledge acquisition. A pioneering work by PICa employs the frozen GPT-3 model to answer the question with a formatted prompt as its input \cite{pica}. Given a testing image-question pair, PICa first translates the image into a caption using an off-the-shelf captioning model. The question, caption, and a few in-context examples are then integrated into a textual prompt that can induce GPT-3 to predict the answer directly. Thanks to the powerful knowledge reasoning ability of \mbox{GPT-3}, PICa achieves significant performance improvements compared to those retrieval-based methods using explicit KBs. 
Inspired by PICa, KAT \cite{kat} and REVIVE \cite{revive} learn KB-augmented VQA models to exploit both the implicit knowledge from LLMs and explicit knowledge from KBs for answer prediction. The synergy of the two knowledge resources brings further improvements to their models. Despite the promising results achieved by these methods, they have not fully activated the capability of the LLMs due to the following limitations:     
\begin{enumerate}[(i)]
	\item The captions cannot cover all the necessary information in the image. Consider the example in Fig. \ref{fig:intro}, the caption ``{a group of people walk in a city square}'' contributes nothing to answering the question ``{what fruit comes from these trees}''. In such situation, the LLM has to make an aimless and biased guess to answer the question. 
	\item LLMs like GPT-3 employ a few-shot learning paradigm that requires a few in-context examples to adapt to new tasks. Therefore, the choice of these examples is critical to model performance. As reported in \cite{pica}, existing example selection strategies achieve far inferior performance to the oracle strategy that selects examples for a testing sample based on the similarity of their ground-truth answers, which is unavailable during testing.
\end{enumerate}
We ask: \emph{Is it possible to endow the LLM with some \textbf{heuristics} to enhance its capacity for knowledge-based VQA?} 

In this paper, we present \textbf{Prophet}---a conceptually simple yet effective framework designed to \textbf{\underline{pro}}m\textbf{\underline{p}}t LLMs with answer \textbf{\underline{he}}uris\textbf{\underline{t}}ics for knowledge-based VQA. By answer \mbox{heuristics}, we mean some promising answers that are presented in a proper manner in the prompt. Specifically, we introduce two types of complementary answer heuristics, namely \emph{answer candidates} and \emph{answer-aware examples}, to overcome the limitations in (i) and (ii), respectively. Given a testing input consisting of an image and a question, the answer candidates refer to a list of promising answers to the testing input, where each answer is associated with a confidence score. The answer-aware examples refer to a list of in-context examples, where each example has a similar answer to the testing input. Fortunately, these two types of answer heuristics can be simultaneously obtained from any vanilla VQA model trained on a specific knowledge-based VQA dataset. A schematic of Prophet is illustrated in \mbox{Fig. \ref{fig:intro}}.

Without bells and whistles, Prophet surpasses previous state-of-the-art single-model results on the challenging OK-VQA and A-OKVQA datasets \cite{okvqa,a-okvqa}, including the heavily-engineered Flamingo-80B model trained on 1.8B image-text pairs \cite{flamingo}. Moreover, Prophet is friendly to most researchers, as our results can be reproduced using a single GPU and a number of GPT-3 invocations. 

A preliminary version of this manuscript was published in \cite{prophet}. Based on that version, we have made the following contributions to further improve the capabilties and generality of Prophet: (i) we investigate diverse types of VQA models, including the classical discriminative models trained from scratch and the latest generative VQA models pretrained on large-scale corpus; (ii) we expand the used LLM from the commercial GPT-3 model to a wide range of open-source models; (iii) apart from OK-VQA and A-OKVQA, we conduct more experiments on two other knowledge-based VQA datasets, namely ScienceQA \cite{scienceqa} and TextVQA \cite{textvqa}. Furthermore, we present an improved Prophet++ framework by introducing large multimodal models (LMMs) into Prophet, which generates a new type of answer heuristic (\emph{i.e.}, \emph{answer-aware rationales}) in the first stage and enables an extra visual input in the second stage.

The source code is made available here\footnote{\url{https://github.com/MILVLG/prophet}}. We hope our studies may inspire future research on knowledge-based VQA and universal vision-language learning.

\begin{figure*}
	\begin{center}
		\includegraphics[width=\linewidth]{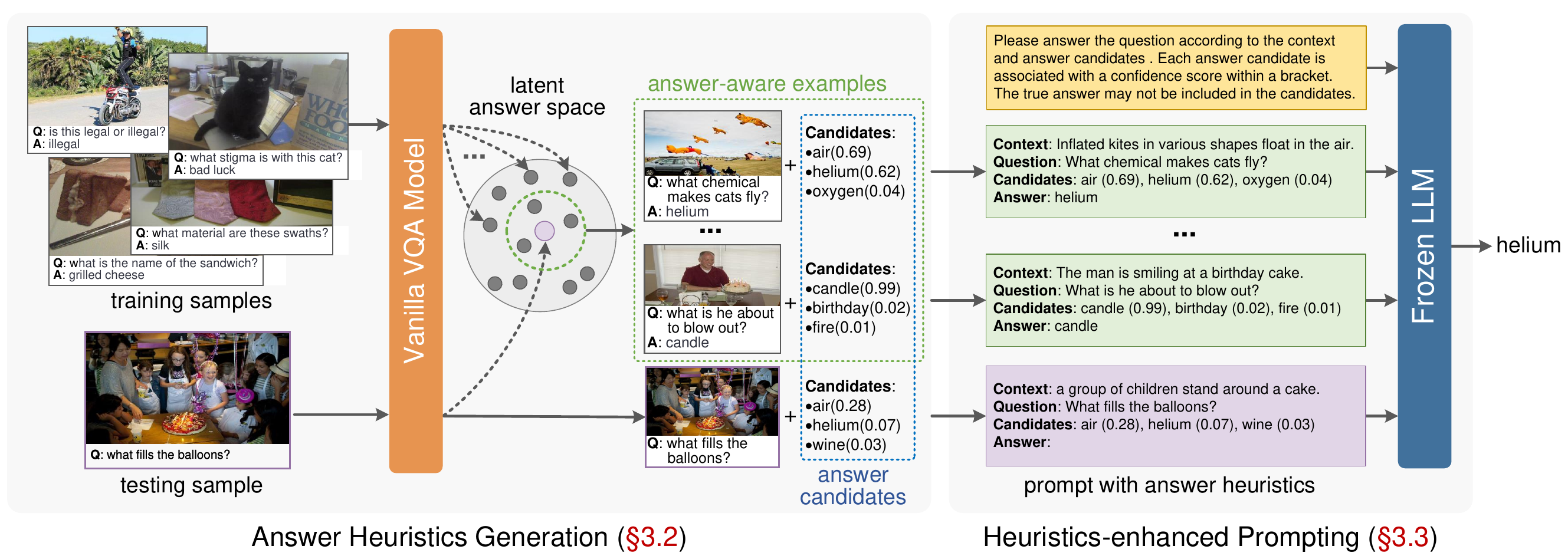}
		\caption{\textbf{Our Prophet framework} has two stages: answer heuristics generation and heuristics-enhanced prompting. In the answer heuristics generation stage, a vanilla VQA model trained on specific knowledge-based VQA dataset is employed to generate two types of complementary answer heuristics, \emph{i.e.}, answer candidates and answer-aware examples. In the heuristics-enhanced prompting stage, the answer heuristics, question, and caption are integrated into a formatted prompt to instruct a frozen LLM (\emph{e.g.}, GPT-3) to predict an answer. As shown in the example, both answer heuristics contribute to the answer of ``helium''.}
		\label{fig:framework}
		\vspace{-15pt}
	\end{center}
\end{figure*}
\section{Related Work}
\noindent\textbf{Visual Question Answering (VQA).} VQA has been of growing interest over the last few years. Recent studies in VQA research can be roughly divided into the following categories: better visual features \cite{butd,vinvl,clip-vil}, more powerful model architectures \cite{ban,regat,mcan}, and more effective learning paradigms \cite{blip,vilbert,lxmert,uniter,cui2021rosita}. Most current state-of-the-art VQA methods employ the Transformer architecture \cite{transformer}. By incorporating vision-language pretraining on large-scale datasets, they have approached or even surpassed human-level performance on several representative benchmarks \cite{flamingo,ofa}. Besides these studies on general-purpose VQA, there is also a growing trend towards exploring more granular VQA tasks with specific reasoning skills, \emph{e.g.}, neural-symbolic reasoning \cite{clevr,gqa} and knowledge utilization \cite{fvqa,okvqa}.  
\vspace{5pt}
\\
\noindent\textbf{Knowledge-based VQA.} The core of this task lies in {knowledge acquisition and integration}. Early explorations parse the inputs into structured queries and retrieve supporting knowledge from fixed knowledge bases (KBs) to obtain the answers \cite{fvqa,kbvqa}. As the provided knowledge resources are not sufficient to represent general knowledge, subsequent research mainly focuses on acquiring explicit knowledge from multiple open-domain knowledge resources, \emph{e.g.}, ConceptNet \cite{conceptnet}, Wikipedia \cite{wikidata}, and Google Images \cite{mavex}. This retrieved knowledge is integrated with the image-question pair for answer prediction \cite{mavex,trig}. Motivated by the powerful capacities of LLMs (\emph{e.g.}, GPT-3 \cite{gpt3}) in knowledge reasoning, recent state-of-the-art approaches regard an LLM as an implicit knowledge engine. They either utilize it to predict answers from given questions and extract visual captions \cite{pica} or to extract answer candidates with evidence to improve answer prediction \cite{kat,revive}. Nevertheless, they have not fully activated the reasoning capability of LLMs, as the necessary visual information to answer the question is not represented exactly. 

This motivates us to explore the strategies for prompting LLMs with question-aware information (\emph{i.e.}, answer heuristics). Similar to Prophet, a concurrent work PromptCap also aims to enhance the input information for LLMs by learning a question-aware captioning model \cite{promptcap}. However, PromptCap needs to use LLM in both the training and testing phases, which incurs tremendous computational costs as the training set is usually large. In contrast, Prophet is more economical as it only utilizes LLM in the testing phase.
\vspace{5pt}
\\
\noindent\textbf{In-context learning.} Unlike the \emph{pretrain-then-finetune} paradigm for language models like BERT \cite{bert}, GPT-3 innovatively introduces a \emph{few-shot in-context learning} paradigm and has become the de facto standard for subsequent LLMs. To adapt to a new task, GPT-3 only needs to concatenate a few examples of the task with the input as the \emph{prompt} at inference time and requires no parameter updates. This appealing property has inspired research on training multimodal few-shot learners \cite{flamingo}. Empirical studies show that a huge model (\emph{e.g.}, 80B parameters in Flamingo \cite{flamingo}) is required for effective few-shot learning, which is unaffordable for most people to reproduce their results.

\section{The Prophet and Prophet++ Frameworks}
As depicted in Fig. \ref{fig:framework}, our Prophet is a conceptually simple two-stage framework. In the answer heuristics generation stage, a vanilla VQA model is learned to generate two types of answer heuristics, \emph{i.e.}, answer candidates and answer-aware examples (detailed in $\S$\ref{sec:ahg}). In the heuristics-enhanced prompting stage, the answer heuristics, question, and caption are integrated into a formatted prompt to instruct a frozen LLM to predict an answer (detailed in $\S$\ref{sec:prompting}). Moreover, we introduce modern large multimodal models (LMMs) in different stages of Prophet to obtain a more powerful Prophet++ framework, which is detailed in $\S$\ref{sec:prophet++}.

\subsection{Preliminaries}\label{sec:pre}
Before presenting the Prophet, we briefly introduce the in-context learning paradigm developed by GPT-3 and its adaptation to knowledge-based VQA by PICa \cite{pica}. 

GPT-3 is an autoregressive language model pretrained on a tremendous dataset. During inference, in-context few-shot learning formulates a new downstream task as a text sequence generation task on the frozen model. Given a testing input $\boldsymbol{x}$, its target $\boldsymbol{y}$ is predicted conditioned on a formatted prompt $\boldsymbol{p}(\boldsymbol{h}, \mathcal{E}, \boldsymbol{x})$, where $\boldsymbol{h}$ refers to a prompt head, \emph{aka} instruction, that describes the task, $\mathcal{E}=\{\boldsymbol{e}_1, ...,\boldsymbol{e}_n\}$ corresponds to $n$ in-context examples. Let the target $\boldsymbol{y}=(y^1,...,y^L)$ be a text sequence of $L$ tokens. For notational convenience, we denote $[l]$ as a set of natural numbers from 1 to $l$ and use $\boldsymbol{y}^{[l]}=(y^1,...,y^{l})$ to represent a sub-sequence containing the first $l$ words of $\boldsymbol{y}$. At each decoding step $l$, we have: 
\begin{equation}
	y^l = \mathop{\text{argmax}}\limits_{\hat{y}^l} p_\textrm{GPT-3}(\hat{y}^l|\boldsymbol{p}, \boldsymbol{y}^{[l-1]})
\end{equation}
where each in-context example $\boldsymbol{e}_i=(\boldsymbol{x}_i,\boldsymbol{y}_i)$ contains an input-target pair of the task, which is constructed manually or sampled from the training set.

To adapt LLMs like GPT-3 to address the knowledge-based VQA task, the key is to design proper prompts. Given a question $q$ and an image $v$ as inputs, the VQA task aims to predict a target answer $a$. Since LLMs do not understand images intrinsically, the image needs to be translated into a caption $c$ using an off-the-shelf captioning model. PICa formulates the testing input $\boldsymbol{x}$ as the following template:
\begin{table}[H]
	\centering
	\begin{tabular}{|l|}
		\hline
		\texttt{Context:} \textcolor{blue}{$c$}\enspace \textbackslash\texttt{n} \enspace
		\texttt{Question:} \textcolor{blue}{$q$}\enspace \textbackslash\texttt{n} \enspace
		\texttt{Answer:} \enspace\enspace\enspace\enspace\enspace\enspace\enspace\enspace\enspace\enspace\enspace\enspace\\
		\hline
	\end{tabular}
\end{table}
\noindent
where the variables marked in \textcolor{blue}{blue} will be substituted by specific testing inputs. \textbackslash\texttt{n} stands for a line break in the template.  Accordingly, each in-context example $\boldsymbol{e_i}$ is formulated into a similar template as follows:
\begin{table}[H]
	\centering
	\begin{tabular}{|l|}
		\hline
		\texttt{Context:} \textcolor{blue}{$c_i$}\enspace \textbackslash\texttt{n} \enspace
		\texttt{Question:} \textcolor{blue}{$q_i$}\enspace \textbackslash\texttt{n} \enspace
		\texttt{Answer:} \textcolor{blue}{$a_i$}\enspace\enspace\enspace\enspace\enspace\enspace\enspace\enspace\enspace\\
		\hline
	\end{tabular}
\end{table}
\noindent
where $c_i$, $q_i$, and $a_i$ refer to an image-question-answer triplet collected from the training set. The complete prompt of PICa consists of a fixed prompt head, a few in-context examples, and a testing input. This prompt is fed into a frozen LLM for answer prediction.

Our Prophet inherits the pipeline of PICa. In addition, we introduce answer heuristics into the prompt structure to better activate the reasoning capability of the LLM, which leads to more accurate answers.

\subsection{Stage-1: Answer Heuristics Generation}\label{sec:ahg}
We introduce two types of answer heuristics: answer candidates and answer-aware examples. Given a testing input consisting of an image and a question, the answer candidates refer to a list of promising answers to the testing input, where each answer is associated with a confidence score. The answer-aware examples refer to a list of in-context examples, where each example has similar answers to the testing input. 
Interestingly, these two types of answer heuristics can be obtained simultaneously from any vanilla VQA model trained on specific knowledge-based VQA task. 

\begin{figure}
	\begin{center}
		\includegraphics[width=\linewidth]{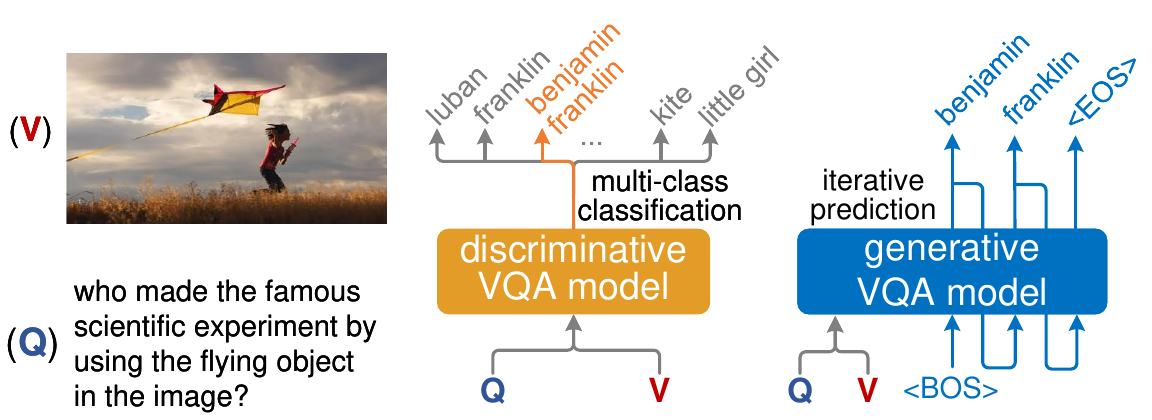}
		\caption{\textbf{Discriminative \emph{vs.} generative VQA models.} Taking an image (V) and a question (Q) as inputs, a typical discriminative VQA model performs multi-class classification to predict the most relevant answer (may contain multiple words) from a predefined answer vocabulary, while a typical generative VQA model iteratively predicts one answer word at a time to constitute the final answer.}
		\label{fig:dis-gen}
		\vspace{-15pt}
	\end{center}
\end{figure}

As shown in Fig. \ref{fig:dis-gen}, existing VQA methods can be categorized into \emph{discriminative} and \emph{generative} ones based on the ways they obtain answers. This discrepancy leads to different strategies for answer heuristics generation. We elaborate the strategy for each class of VQA models below. 

\subsubsection{Discriminative VQA models}
Denote a VQA training dataset as $\mathcal{D}=\{(v_i, q_i, a_i)\}_{i=1}^M$, where $v_i, q_i, a_i$ refer to the image, question, and answer, respectively. The most frequent answers in the training set form an answer vocabulary $\mathcal{V} = \{\boldsymbol{w}_j\}_{j=1}^S$, where $S$ is the answer vocabulary size. A discriminative VQA model $\mathcal{M}_\mathrm{disc}$ is learned from $\mathcal{D}$ to perform an $S$-way classification over the answers. Generally, the model $\mathcal{M}_\mathrm{disc}$ can be separated into two submodels, \emph{i.e.}, a backbone $\mathcal{M}^{B}_\mathrm{disc}$ and a prediction head $\mathcal{M}^{H}_\mathrm{disc}$. The backbone $\mathcal{M}^{B}_\mathrm{disc}$ acts as an encoder to fuse multimodal inputs $v$ and $q$ and obtain a fused feature $z$: 
\begin{equation}\label{eq:mback}
	{z} = \mathcal{M}^{B}_\mathrm{disc}(v, q)
\end{equation}
The prediction head $\mathcal{M}_H$ simply adopts a linear layer followed by a sigmoid function to project the fused feature $z$ into a score vector $y\in\mathbb{R}^S$ over the answer vocabulary:
\begin{equation}\label{eq:mhead}
	{y} = \mathcal{M}^{H}_\mathrm{disc}({z})
\end{equation}
where the $i$-th element of $y$ represents the confidence score for answer $\boldsymbol{w}_i$. Based on the above definitions, we explain how to generate the two types of answer heuristics below.
Note that although the learned VQA model $\mathcal{M}_\mathrm{disc}$ does not incorporate any external knowledge, it can be used for knowledge-based VQA when trained properly. We regard it as a reference model and compare its performance to Prophet in the experiments to show the effectiveness of LLM for knowledge-based VQA.
\vspace{5pt}
\\
\noindent\textbf{Answer candidates.} Given a testing input $(v,q)$, we obtain its score vector ${y}$ for all answers using Eq.(\ref{eq:mhead}). Denoting $s_{i}\in\mathbb{R}^+$ as the $i$-th element of $y$, we obtain the top-$K$ answers with the highest scores as follows:
\begin{equation}
	\mathcal{I}_\textrm{AC} = \mathop{\text{argTopK}}\limits_{j\in\{1,2,...,S\}}s_{j}
\end{equation}
where $\mathcal{I}_\textrm{AC}$ denotes an index set of the top-$K$ answer candidates. The answer candidates $\mathcal{C}$ are defined as follows: 
\begin{equation}\label{eq:ac_disc}
	\mathcal{C} = \{(\boldsymbol{w}_{j},s_j)\ |\ j\in \mathcal{I}_\textrm{AC}\}
\end{equation}
where $\boldsymbol{w}_{j}$ and $s_j$ are an answer candidate and its confidence score, respectively. To make the formats of the in-context examples and testing input consistent, for each example $\boldsymbol{e_i}$ we also calculate and provide a set of answer candidates $\mathcal{C}_i$.
\vspace{5pt}
\\
\noindent\textbf{Answer-aware examples.} 
Several previous studies have shown that the choice of in-context examples is crucial for LLM's few-shot learning \cite{pica}. Their results motivate us to devise an \emph{answer-aware} example selection strategy. 

Given a testing input $(v,q)$ and any training input $(v_i,q_i)$, we can obtain their corresponding fused features $z$ and $z_i$ from Eq.(\ref{eq:mback}) using the trained model. Since the fused features are linearly projected for answer prediction, we conjecture that these fused features lie in a \emph{latent answer space} that contains rich semantics of the answers to the given image-question pairs. If $z$ and $z_i$ are close in the latent space, they are more likely to share similar answers and image-question inputs. 

We calculate the cosine similarity of the fused feature between the testing input and each training input, then select top-$N$ nearest neighbors in the latent space as the answer-aware examples: 

\begin{equation}
	\mathcal{I}_\textrm{AE} = \mathop{\text{argTopN}}\limits_{i\in\{1,2,...,M\}} \frac{{z}^T{z}_i}{\|{z}\|_2\|{z}_i\|_2}
\end{equation}
where $\mathcal{I}_\textrm{AE}$ is an index set of the top-$N$ similar samples in $\mathcal{D}$. The answer-aware examples $\mathcal{E}$ are defined as follows:
\begin{equation}
	\mathcal{E} = \{(v_{i}, q_{i}, a_{i})\ |\ i\in\mathcal{I}_\textrm{AE}\}
\end{equation}

Note that the fused features of the training inputs can be computed and stored beforehand, allowing efficient answer-aware example selection. 

\subsubsection{Generative VQA models}\label{sec:generative_vqa}
Recent state-of-the-art VQA models tend to use generative model architectures due to their remarkable scalability and generalizability \cite{ofa,blip,mplug}. 

Given the same VQA training dataset $\mathcal{D}=\{(v_i, q_i, a_i)\}_{i=1}^M$ as above, a generative VQA model $\mathcal{M}_\mathrm{gen}$ is learned from $\mathcal{D}$ to generate answers word-by-word from a pre-defined word vocabulary $\mathcal{V}=\{w_j\}_{j=1}^S$, where $S$ is the word vocabulary size. Each answer can be represented as a text sequence with a dynamic length of $L$ words: 
\begin{equation}\label{eq:ans_word}
	\boldsymbol{w}=(w^1,w^2,...,w^L)
\end{equation}
where $w^1=[\mathtt{BOS}]$ refers to a special start-of-sentence token and $w^L=[\mathtt{EOS}]$ refers to an end-of-sentence token.

Similar to the discriminative model, $\mathcal{M}_\mathrm{gen}$ can also be separated into a backbone $\mathcal{M}^B_\mathrm{gen}$ and a prediction head $\mathcal{M}^H_\mathrm{gen}$. The backbone $\mathcal{M}^B_\mathrm{gen}$ corresponds to an encoder-decoder or a pure decoder architecture that fuses multimodal inputs $v$ and $q$, and then generates latent feature of each answer word using an autoregressive manner: 	
\begin{equation}\label{eq:mback_gen}
z^{l}=\mathcal{M}^B_\mathrm{gen}(v,q,\boldsymbol{w}^{[l-1]}) 
\end{equation}
where $z^{l}$ denotes the latent feature of $l$-th answer word.
On top of the latent feature $z^{l}$, the prediction head $\mathcal{M}^H_\mathrm{gen}$ applies a linear projection (or a MLP) followed by a softmax function to decode it into a score distribution $y^{l}\in\mathbb{R}^S$ over the whole word vocabulary:
\begin{equation}\label{eq:mhead_gen}
	y^l=\mathcal{M}^H_\mathrm{gen}(z^l) 
\end{equation}
where the $l$-th answer word $w^l$ is obtained from $y^l$ by greedily choosing the word with the highest score. Until an $[\mathtt{EOS}]$ token is generated, $w^l$ is appended to $\boldsymbol{w}^{[l-1]}$ to obtain $\boldsymbol{w}^{[l]}$, which is iteratively fed into the model $\mathcal{M}_\mathrm{gen}$ to predict the next word.
\vspace{5pt}
\\
\noindent\textbf{Answer candidates.}  Given a testing input $(v,q)$, we can obtain its most relevant answer using the greedy decoding strategy above. However, how to obtain the answer candidates consisting of the top-$K$ answers and their confidence scores is not straightforward. We resort to the \emph{beam search} algorithm, which is widely used in neural machine translation \cite{seq2seq} and visual captioning \cite{mvmt}, to address the issue.

Similar to Eq. (\ref{eq:ac_disc}), we denote the top-$K$ answer candidates as a set of tuples as follows: 
\begin{equation}
	\mathcal{C}=\{(\boldsymbol{w}_1, s_1), (\boldsymbol{w}_2,s_2),...,(\boldsymbol{w}_K,s_K)\}
\end{equation}
where each $\boldsymbol{w}_j$ represents an answer consisting of a sequence of answer words and $s_j\in\mathbb{R}^+$ denotes its corresponding confidence score calculated over all the answer words. The answer candidate set $\mathcal{C}$ is obtained from the generative model $\mathcal{M}_\mathrm{gen}$ equipped with the beam search strategy.  

Specifically, we initialize each answer $\boldsymbol{w}_j$ with the same $[\mathtt{BOS}]$ token. At each decoding step $l$, each $\boldsymbol{w}_j$ of length $l$ is first passed through $\mathcal{M}_\mathrm{gen}$ to obtain its top-$K$ candidate words with the highest scores. After that, an \emph{expand-then-reduce} strategy is performed to update the $K$ answers: (i) \textbf{expand step}: each $\boldsymbol{w}_j$ is expanded $K$ times to combine with the $K$ candidate words, resulting in $K*K$ new candidates answers of length $l+1$; (ii) \textbf{reduce step}: among the $K*K$ candidate answers, only the top-$K$ ones with the highest accumulated scores $s=\sum_{i=1}^l\mathrm{log}~y^i$ are retained, which are then regarded as the inputs to the next decoding step.
\vspace{5pt}
\\
\noindent\textbf{Answer-aware examples.} Similar to the example selection strategy for discriminative models, the answer-aware examples for generative models are also obtained by performing kNN search in a latent answer space. 
It is worth noting that the granularity of the latent features is different for the two types of VQA models: each latent feature obtained from a discriminative VQA model refers to an answer entry in the answer vocabulary, while each latent feature obtained from a generative VQA model refers to an answer word. 

Given a testing input $(v,q)$ and $i$-th training input $(v_i, q_i)$, the latent features for their multi-word answers can be respectively represented as feature groups $Z=[z^1,z^2,...,z^L]\in\mathbb{R}^{L\times d}$ and $Z_i=[z_i^1,z_i^2,...,z_i^{L_i}]\in\mathbb{R}^{L_i\times d}$, where $d$ is the common dimensionality of the latent answer space, $L$ and $L_i$ refer to the answer lengths of $Z$ and $Z_i$, respectively. We define a simple score function as follows to average the dot-product similarity of each paired features $z_j\in Z$ and $z_i^k\in Z_i$ :
\begin{equation}\label{eq:score_func}
\pi_i=\frac{1}{L*L_i}\sum\limits_{j=1}^L\sum\limits_{k=1}^{L_i} \frac{z^jz_i^k}{\|z^j\|_2\|z_i^k\|_2}
\end{equation}

Using Eq. (\ref{eq:score_func}), we obtain the top-$N$ nearest neighbors of the query input in the training set and then format them as the answer-aware examples $\mathcal{E}$ as follows:
\begin{equation}
	\begin{aligned}
	\mathcal{I}_\textrm{AE} &= \mathop{\text{argTopN}}\limits_{i\in\{1,2,...,M\}}\pi_i\\
	\mathcal{E} &= \{(v_{i}, q_{i}, a_{i})\ |\ i\in\mathcal{I}_\textrm{AE}\}
	\end{aligned}
\end{equation}
where $\mathcal{I}_\textrm{AE}$ is an index set of the top-$N$ nearest neighbors in the training set $\mathcal{D}$.
   
\subsection{Stage-2: Heuristics-enhanced Prompting} \label{sec:prompting}
After obtaining the answer heuristics (\emph{i.e.}, answer candidates $\mathcal{C}$ and answer-aware examples $\mathcal{E}$) from the stage-1, we encode them into a heuristics-enhanced prompt to facilitate the few-shot learning capacity of the LLM for knowledge-based VQA.

A prompt consists of a prompt head, a set of in-context examples, and a testing input. The prompt head describes the VQA task in natural language. We refer to the prompt head designed in PICa and supplement it with a new description of the answer candidates. Although we encourage LLM to generate answers according to the answer candidates, we also allow it to take broad explorations and generate answers beyond the candidates. The complete format of our prompt head is shown in the yellow box of Fig. \ref{fig:framework}.

Our in-context examples are derived from the obtained $N$ answer-aware examples $\mathcal{E}=\{\boldsymbol{e}_1,\boldsymbol{e}_2,...,\boldsymbol{e}_N\}$. Based on PICa's template in $\S$\ref{sec:pre}, for example $\boldsymbol{e}_i$, we introduce its answer candidates $\mathcal{C}_i$ by adding \emph{one} line of code as follows:

\begin{table}[H]
	\centering
	\begin{tabular}{|l|}
		\hline
		\texttt{Context:} \textcolor{blue}{$c_i$} \enspace \textbackslash\texttt{n} \enspace
		\texttt{Question:} \textcolor{blue}{$q_i$} \enspace \textbackslash\texttt{n} \\
		\texttt{Candidates:} \textcolor{blue}{$\boldsymbol{w}_{j_1}$}(\textcolor{blue}{$s_{j_1}$}), \textcolor{blue}{$\boldsymbol{w}_{j_2}$}(\textcolor{blue}{$s_{j_2}$}),...,\textcolor{blue}{$\boldsymbol{w}_{j_K}$}(\textcolor{blue}{$s_{j_K}$})\enspace \textbackslash\texttt{n} \enspace\enspace\enspace\enspace\enspace\enspace\enspace\enspace\enspace\enspace\enspace\\ 
		\texttt{Answer:} \textcolor{blue}{$a_i$}\\
		\hline
	\end{tabular}
	\vspace{-5pt}
\end{table}
\noindent
where $j_1, j_2, \cdots, j_K$ correspond to the actual indices of the elements in $\mathcal{C}_i$.
Each answer candidate $w_{j_k}$ is paired with its confidence score $s_{j_k}$ within a bracket. The confidence scores additionally offer the reliability of the corresponding answer candidates, which helps the LLM focus more on the promising candidates and be more tolerant of the less relevant candidates. For the testing input, its template is similar to that for the in-context examples, except that the answer slot is left blank for the LLM to fill with.

To better exploit available examples, we use the multi-query ensemble strategy \cite{pica}. Specifically, we increase the number of answer-aware examples to $N$*$T$ to obtain $T$ paralleled prompts, where each prompt still contains $N$ examples. By prompting the LLM for $T$ times, we obtain $T$ answer predictions. The majority voting is performed over the $T$ predictions to determine the final answer. The effects of different $N$ and $T$ will be verified in the experiments. 

\subsection{Prophet++ with Frozen Large Multimodal Models}\label{sec:prophet++}
\begin{figure}
	\begin{center}
		\includegraphics[width=\linewidth]{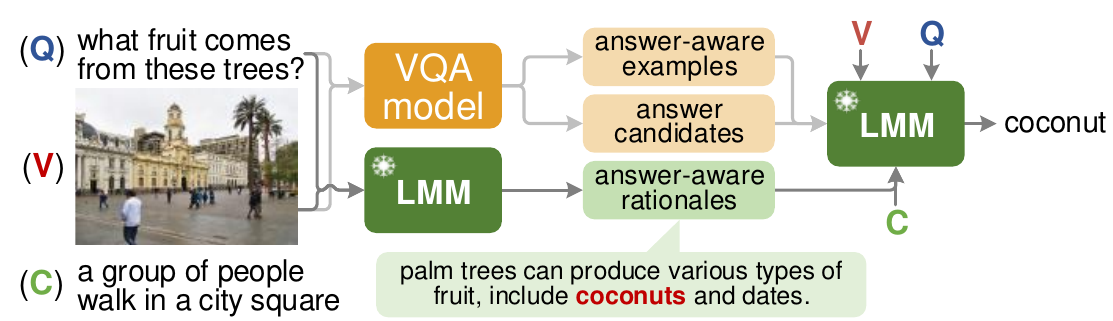}
		\caption{\textbf{The Prophet++ framework} additionally introduces large multimodal models (LMMs) in different stages of Prophet. Specifically, the LMM in stage-1 is used to generate a new type of answer heuristic, \emph{i.e.}, the \emph{answer-aware rationales} while the LMM in stage-2 is used to handle an extra visual input (V), thus providing more comprehensive knowledge to answer the question.}
		\label{fig:prophet++}
		\vspace{-10pt}
	\end{center}
\end{figure}

The original design of Prophet is based on pure-text LLMs that cannot perceive the images. Recently, a series of large multimodal models (LMMs), \emph{e.g.}, GPT-4V, GPT-4o, and LLaVA \cite{llava}, have been proposed and show remarkable capabilities on various multimodal tasks. This raises a natural question: ``\emph{Can these powerful LMMs be utilized in Prophet to further facilitate its capabilities?}'' To this end, we propose Prophet++, an extended framework that utilizes LMMs in both stages of Prophet, as shown in Fig \ref{fig:prophet++}. 

In the first stage of Prophet++, we first use the VQA model to extract the two aforementioned answer heuristics in Prophet. Inspired by the strong reasoning capability of the latest LMMs, we extract a new type of answer heuristic, \emph{i.e.}, answer-aware rationales, which contains the reasoning process to answer the question. The prompt to generate answer-aware rationales is as follows:
\begin{table}[H]
	\centering
	\begin{tabular}{|l|}
		\hline
		\texttt{You are provided with an image and a question. Please~~}\\ \texttt{think step by step to generate the rationale to help~~~}\\ \texttt{answering the question.}\\
		\hline
	\end{tabular}
	\vspace{-5pt}
\end{table}
where a few examples are also provided to control the output format. The output rationale is appended to the extracted image caption to be used in the second stage.   

In the second stage of Prophet++, we replace the frozen LLM in Prophet with an LMM to handle multimodal inputs. Note that it takes a considerable amount of tokens to encode an image by a typical LMM. To restrict the total length of the multimodal inputs, only the test image is fed into the LMM while the images of the in-context examples are omitted.

\section{Experiments}
We mainly evaluate the performance of Prophet on two prevalent knowledge-based VQA datasets: OK-VQA \cite{okvqa} and A-OKVQA \cite{a-okvqa}. We conduct comprehensive ablation experiments to explore the effectiveness of Prophet. By taking the ablation results into account, we perform thorough comparisons of Prophet and state-of-the-art methods. Moreover, we showcase the generalization ability of Prophet on two diverse knowledge-based VQA datasets ScienceQA \cite{scienceqa} and Text-VQA \cite{textvqa}, which require external science and OCR knowledge, respectively.

\subsection{Datasets}
\noindent\textbf{OK-VQA} is a commonly used knowledge-based VQA dataset \cite{okvqa}. The dataset contains 9K and 5K image-question pairs for training and testing, respectively. All questions are manually filtered to ensure that outside knowledge is required to answer the questions. 
\vspace{5pt}
\\
\noindent
\textbf{A-OKVQA} is currently the largest knowledge-based VQA dataset \cite{a-okvqa}. The dataset is split into three subsets: 17K training, 1K validation, and 7K testing. 
Both direct answering (DA) and multiple choice (MC) evaluations are provided. 
\vspace{5pt}
\\
\noindent
\textbf{ScienceQA} is a dataset that consists of about 21K questions over a diverse set of high school-level science topics \cite{scienceqa}. Out of the 21K questions, only the `IMG' subset of 10.3K (48.7\%) samples with images are used in our experiments. 
\vspace{5pt}
\\
\noindent
\textbf{TextVQA} contains 28K images and 45K questions, where each question requires models to read and reason about the text in the image \cite{textvqa}. The dataset is split into three subsets of 34.6K training, 5K validation, and 5.7K testing questions. Following \cite{tap,latr}, we supplement the training set with the augmented VQA samples from ST-VQA \cite{stvqa}.
\begin{table*}
	\renewcommand\arraystretch{1.2}
	\scriptsize
	\begin{subtable}[t]{0.35\textwidth}
		\centering
		\begin{tabular}{lcc}
			VQA model, paradigm & stage-1 acc. & accuracy \\
			\ChangeRT{1.2pt}
			ViLBERT, retrieval \cite{mavex} & 35.20 & 40.28 (+5.08)\\
			ViLBERT, prompt$^\dag$ & 35.28 & 44.97 (+9.69) \\
			\\
			\\
		\end{tabular}
		\captionsetup{font=footnotesize}
		\subcaption{\textbf{Prompting \emph{vs}. retrieval.} Our prompting-based paradigm is more effective than the retrieval-based one in MAVEx \cite{mavex}. $^\dag$: our re-implementation.}
		\label{table:mavex}
	\end{subtable}
	\quad
	\renewcommand\arraystretch{1.2}
	\scriptsize
	\begin{subtable}[t]{0.35\textwidth}
		\centering
		\begin{tabular}{lcc}
			visual features & stage-1 acc. & accuracy \\
			\ChangeRT{1.2pt}
			Bottom-Up \cite{butd} & 46.83 & 55.34 (+8.51) \\
			VinVL \cite{vinvl} & 47.88 & 56.23 (+8.35) \\
			CLIP-ViT-L/14 \cite{clip} & 52.03 & 60.12 (+8.09) \\
			CLIP-RN50$\times$64 \cite{clip} & \textbf{53.04}& \multicolumn{1}{>{\columncolor{mygray}}c}{\textbf{60.84 (+7.80)}} \\
		\end{tabular}
		\captionsetup{font=footnotesize}
		\subcaption{\textbf{Capability of VQA models.} More powerful VQA models lead to higher accuracies, but obtain slightly less relative improvements from stage-2. }
		\label{table:vqa_model}
	\end{subtable}
	\quad
	\renewcommand\arraystretch{1.2}
	\scriptsize
	\begin{subtable}[t]{0.26\textwidth}
		\centering
	\begin{tabular}{ccc}
		\#candidates ($K$) & hit rate & accuracy \\
		\ChangeRT{1.2pt}
		0 & - & 49.63 \\ 
		1 & 53.04 & 56.04 \\ 
		5 & 75.20 & 60.17 \\ 
		10 & \textbf{79.83}& \multicolumn{1}{>{\columncolor{mygray}}c}{\textbf{60.84}}\\  
	\end{tabular}
	\captionsetup{font=footnotesize}
	\subcaption{\textbf{Answer candidates.} They are crucial to Prophet and increasing $K$ leads to better performance.}
	\label{table:noc}
\end{subtable}
\\[1.em]
\renewcommand\arraystretch{1.2}
\scriptsize
\begin{subtable}[t]{0.35\textwidth}
	\centering
	\begin{tabular}{lcc}
		example selection & ~~~hit rate~~~& accuracy \\
		\ChangeRT{1.2pt}
		(a) rand  & 5.31  & 58.66 \\
		(b) ques + img \cite{pica} & 59.58  & 59.82\\
		(c) fused & \textbf{83.63} & \multicolumn{1}{>{\columncolor{mygray}}c}{\textbf{60.84}}\\
		(d) fused + ques + img ~~~~~~~& 82.45 & 60.38 \\
		(e) answer logits & 79.25 & 60.40 \\
	\end{tabular}
	\captionsetup{font=footnotesize}
	\subcaption{\textbf{Example selection strategy.} Our answer-aware example selection based on fused features is more effective than the others.}
	\label{table:example}
\end{subtable}
\quad
\renewcommand\arraystretch{1.2}
\scriptsize
\begin{subtable}[t]{0.35\textwidth}
	\centering
	\begin{tabular}{ccc}
		\#examples ($N$) & accuracy ($T$=1) & accuracy ($T$=5) \\
		\ChangeRT{1.2pt}
		0 & 49.97 & 49.97 \\
		1 & 54.89 & 56.75 \\
		8 & 57.49 & 59.91 \\
		16 & 57.52 &\multicolumn{1}{>{\columncolor{mygray}}c}{60.84} \\
		20 & 57.91 & \textbf{61.10}\\
	\end{tabular}
	\captionsetup{font=footnotesize}
	\subcaption{\textbf{Numbers of examples and queries.} Increasing $N$ and $T$ improves model performance at the expense of linearly increasing overheads.}
	\label{table:noe}
\end{subtable}
\quad
\renewcommand\arraystretch{1.2}
\scriptsize
\begin{subtable}[t]{0.26\textwidth}
	\centering
	\begin{tabular}{lc}
		variants & accuracy \\
		\ChangeRT{1.2pt}
		(a) default & \multicolumn{1}{>{\columncolor{mygray}}c}{\textbf{60.84}} \\
		(b) w/o prompt head & 60.54 \\
		(c) w/o confidence scores & 55.46 \\
		(d) w/o image captions & 58.27 \\
		(e) default+tags \cite{pica} & 60.51 \\
	\end{tabular}
	\captionsetup{font=footnotesize}
	\subcaption{\textbf{Prompt contents.} The default settings contain the exact necessary information for prompting.}
	\label{table:other}
\end{subtable}
\caption{\textbf{Ablation experiments for Prophet}. All the reported results are evaluated on the testing set of OK-VQA v1.1. The best result in each table is bolded and the result with the default settings is marked in \colorbox{mygray}{gray}.}
\label{table:abla}
\end{table*}

\subsection{Implementation Details} \label{sec:implementation}
\noindent
\textbf{Default settings on OK-VQA.} We use the MCAN-large \cite{mcan} as our default VQA model to generate answer heuristics. To improve the model capability, we modify the original MCAN model by: (i) replacing the original bottom-up-attention region-based features with the grid-based features extracted from CLIP's visual encoder with a RN50$\times$64 backbone \cite{clip}; and (ii) replacing the original LSTM network with a pretrained BERT-large model \cite{bert}.  

Similar to \cite{krisp}, we apply the transfer learning paradigm to further enhance the model capability. The model is first pretrained on the VQAv2 dataset \cite{vqa2} and Visual Genome dataset \cite{visualgenome}. To prevent data contamination, we remove those samples from the pretraining dataset, whose images are used in the testing split of OK-VQA. After that, the pretrained model is further finetuned on the training split of OK-VQA to obtain our final VQA model. 
Note that the answer vocabulary of the pretrained model (with 3,129 answers) is quite different from the vocabulary of OK-VQA. To bridge this gap, we merge the answer vocabulary of OK-VQA\footnote{Similar to \cite{butd}, we collect answers that appear more than eight times in the training set of OK-VQA, resulting in 2,794 answers.} with the existing vocabulary, resulting in an expanded answer vocabulary with 4,477 answers for model finetuning. This model is trained on a \emph{single} Nvidia RTX 3090 GPU, which is affordable for most people.

During the prompting stage using LLMs, we follow PICa to use OSCAR+ as the captioning model \cite{vinvl}. Unless otherwise noted, we set the number of answer candidates $K$=10, the number of in-context examples $N$=16, and the number of queries $T$=5 as our default settings. The default version of GPT-3 used in our experiments is \texttt{text-davinci-002} and the sampling temperature is set to 0.

\vspace{5pt}
\noindent
\textbf{Settings on other datasets.}
The settings and strategies for OK-VQA can be directly transferred to A-OKVQA to address its DA task. For the MC task, we follow the strategy in \cite{a-okvqa} to project the predicted answer to the nearest answer choice. Moreover, we design a Prophet variant for the MC task. It uses a slightly different prompt by injecting the choices to in-context examples and testing input, and instructs the LLM to \emph{choose} the correct one from four choices. 

For ScienceQA, we reuse all the default settings for OK-VQA. If a training sample provides extra textual hint, we simply append the text to the generated caption as the new context of the corresponding image. For TextVQA, we use the commercial system from Amazon to extract OCR from images \footnote{\url{https://aws.amazon.com/textract/}}, whose effectiveness has been verified in previous work \cite{latr}. The extracted OCR texts are provided in both the in-context examples and testing input to instruct the LLM.

\vspace{5pt}
\noindent
\textbf{Settings for other Prophet and Prophet++ variants.} In addition to MCAN, we also experiment on Prophet with a generative VQA model mPLUG \cite{mplug}, which is first pretrained on massive image-text pairs and then finetuned on specific VQA dataset. Following the aforementioned two-stage transfer learning paradigm for MCAN, the pretrained mPLUG model is first finetuned on the VQAv2 dataset and then further finetuned on specific knowledge-based VQA dataset. For Prophet++, we use the state-of-the-art LMM \texttt{GPT-4o} in both stages.

\subsection{Ablation Studies} \label{sec:ablation}
\begin{figure}
	\centering
	\begin{subfigure}[h]{0.4\columnwidth}
		\includegraphics[width=\linewidth]{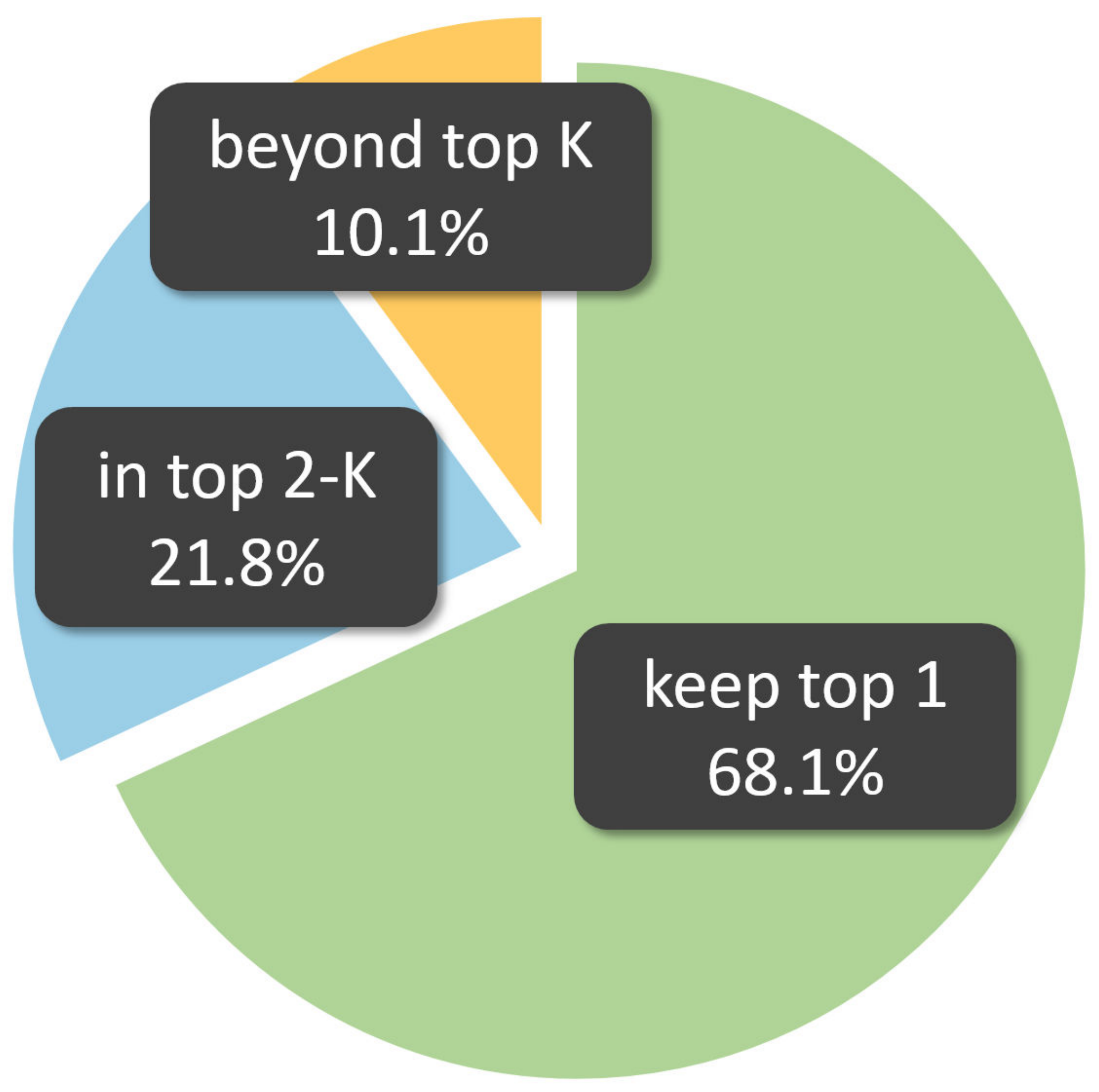}
		\vspace{-9pt}
		\captionsetup{font=footnotesize}
		\caption{behavior distribution}\label{fig:pie}
	\end{subfigure}
	\begin{subfigure}[h]{.55\columnwidth}
		\includegraphics[width=\linewidth]{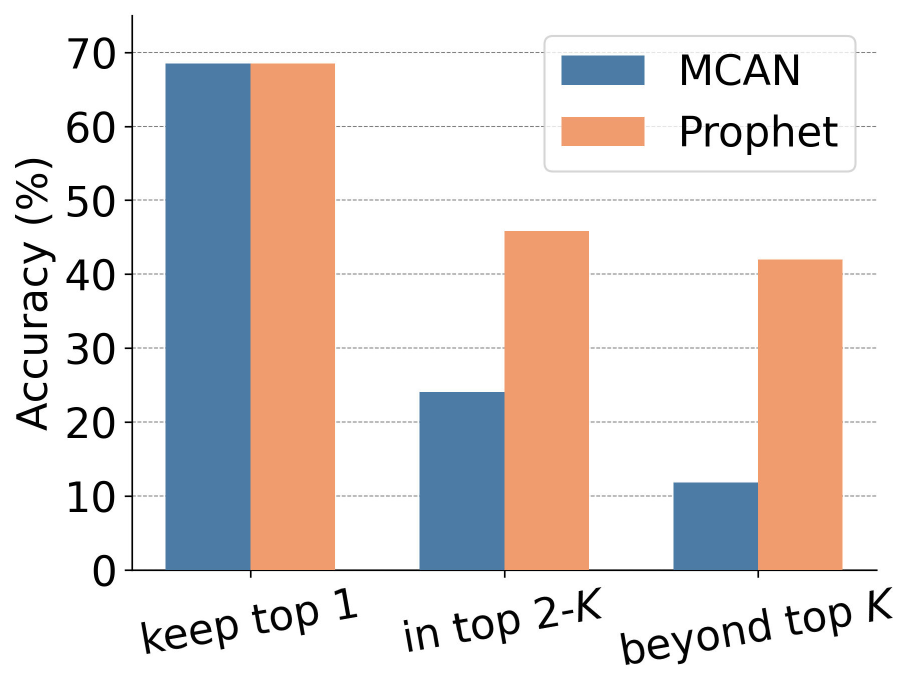}
		\captionsetup{font=footnotesize}
		\caption{per-type accuracy}\label{fig:bar}
	\end{subfigure}
	\caption{\textbf{Prophet's prediction behaviors} in terms of (a) distribution and (b) per-type accuracy. As Prophet takes $K$ answer candidates from MCAN as inputs, we define three prediction behaviors of Prophet, namely ``keep top 1'', ``in top 2-$K$'', and ``beyond top $K$'' predictions of MCAN, respectively. Note that all the testing samples can be categorized into one of the three classes above. }
	\label{fig:stat}
	\vspace{-5pt}
\end{figure}

We conduct ablation experiments for Prophet and Prophet++ on OK-VQA using the default settings above if not mentioned otherwise. Results shown in Table \ref{table:abla} and Fig. \ref{fig:stat} are discussed in detail below.
\vspace{5pt}
\\
\noindent\textbf{Prompting \emph{vs}. retrieval.} Prophet uses a prompting-based paradigm to predict the answer based on a set of promising answer candidates. In contrast, a previous work MAVEx \cite{mavex} exploits answer candidates but adopts a retrieval-based paradigm to search knowledge from external KBs to determine the answer. As both Prophet and MAVEx train a VQA model to generate answer candidates (stage-1), we can compare the superiority of the two paradigms (stage-2). In Table \ref{table:mavex}, we show the performance of the two paradigms in terms of stage-1 accuracy and final accuracy, respectively.

For a fair comparison, we re-implement the VQA model used in MAVEx, \emph{i.e.}, ViLBERT \cite{vilbert}, to generate answer heuristics for our Prophet. From the results, we can see that based on the same VQA model, our Prophet outperforms MAVEx by a large margin (44.97\% \emph{vs}. 40.28\%), showing the superiority of our prompting-based paradigm over MAVEx's retrieval-based paradigm in external knowledge acquisition and integration.
\vspace{5pt}
\\
\noindent\textbf{Capability of VQA models.} In Table \ref{table:vqa_model}, we study how the VQA models of different capabilities impact the performance of Prophet. To better control the model capability, we use the same MCAN model trained with four visual features: region-based Bottom-Up \cite{butd} and VinVL \cite{vinvl} features and grid-based CLIP features from two backbones (ViT-L/14 and RN50$\times$64) \cite{clip}. Results show that more powerful VQA models (reflected in the stage-1 accuracies) lead to better performance of Prophet, as they provide answer heuristics of higher quality. Combining the results in Table \ref{table:mavex}, we also observe that more powerful VQA models achieve less relative improvements from \mbox{GPT-3}, which can be explained by the intrinsic diminishing return property. As a by-product, we verify that the visual features are important to the performance of knowledge-based VQA, which is consistent with the observations in \cite{revive}. The models with CLIP-based visual features significantly outperform those with region-based features, indicating that the CLIP's visual features contain richer visual knowledge due to large-scale pretraining.

In addition to using different visual features for MCAN, we can also replace the whole MCAN model with any generative models pretrained on large-scale multimodal datasets as mentioned in $\S$\ref{sec:pre}. These results will be reported in the main results.
\vspace{5pt}
\\
\noindent\textbf{Answer candidates.} Table \ref{table:noc} varies the number of answer candidates $K$ from 0 to 10 to explore its effect on Prophet. For each testing sample, if the ground-truth answer is hit by one of the $K$ answer candidates, we accumulate the soft score of that ground-truth answer\footnote{In practice, multiple ground-truth answers are provided. If multiple answers are hit simultaneously, we choose the answer with the largest soft score for accumulation.}. The hit rate is calculated over the testing set by dividing the accumulated score by the number of samples.

From the results, we can see that: (i) without any answer candidates, Prophet's accuracy drops by 6.4 points ($K$=0 \emph{vs}. $K$=1), showing the importance of answer candidates in Prophet; (ii) with the increase of answer candidates, the hit rate and final accuracy grow accordingly but they exhibit a tendency to saturate. This is because the quality of answer candidates eventually meets saturation as $K$ increases; (iii) when $K$=1, the final accuracy is even higher than the hit rate (56.04\% \emph{vs}. 53.04\%), which implies that GPT-3 has a strong capability to correct the wrong answer candidates while keeping the correct ones. 
\vspace{5pt}
\\
\noindent\textbf{Example selection strategy.} To show the effectiveness of our answer-aware example selection strategy, we compare it to other example selection strategies in Table \ref{table:example}. The compared strategies include: (a) \emph{rand}: examples that are randomly selected; (b) \emph{ques + img}: examples that are selected based on the joint similarity of question and image features, which is used in PICa; (c) \emph{fused}: our default strategy that selects examples based on the similarity of fused features; (d) \emph{fused + ques + img}: a combination of our default strategy and PICa's strategy; and (e) \emph{answer logits}: examples that are selected based on the similarity of answer logits obtained in Eq.(\ref{eq:mhead}). Besides the final accuracy, we also report the hit rate of answers within the selected examples for each strategy. 

The results show that the accuracy is positively correlated with the hit rate of answers, which verifies our hypothesis that answer-aware examples contribute significantly to the performance of Prophet. Compared with other strategies, our default strategy (c) achieves the best performance with the highest hit rate. The strategy (d) that integrates other information (ques + img) into the (c) leads to worse performance due to the introduction of irrelevant and noisy information. Finally, strategy (e) reports slightly worse performance than (c). We conjecture that this is because the answer logits have lost too much information of the input question and image, which is also useful for GPT-3 to perform knowledge reasoning.
\vspace{5pt}
\\
\noindent\textbf{Numbers of examples and queries.} Table \ref{table:example} contains the ablation studies for the numbers of examples and queries. We choose different numbers of examples $N\in\{0, 1, 8, 16, 20\}$ for each query and different numbers of queries $T\in\{1,5\}$, respectively. The results show that the performance of Prophet improves with the increase of $N$ and $T$, which is consistent with the results in PICa. By increasing $T$ from 1 to 5, the entries with larger $N$ enjoy greater performance improvements at the expense of linearly increasing overheads. 

Interestingly, the Prophet variant with $N$=0 delivers worse performance than the VQA model in stage-1 (49.97\% \emph{vs}. 53.04\%), even though answer candidates are provided. Meanwhile, when given one example ($N$=1), the Prophet variant distinctly surpasses the VQA model (56.75\% \emph{vs}. 53.04\%). This suggests the necessity of few-shot in-context examples for GPT-3 to activate its capability to adapt to the knowledge-based VQA task.
\vspace{5pt}
\\
\noindent\textbf{Prompt contents.} In Table \ref{table:other}, we ablate the \mbox{prompt} contents in the default settings by: (b) removing the \mbox{prompt} head; (c) removing the confidence scores for answer candidates; (d) removing image captions; and (e) adding predicted tags from external models \cite{pica}.
\captionsetup[subfigure]{font=small}

The results lead to the following observations: First, the confidence scores are of critical importance to the performance of our Prophet. This is because they carry the necessary information for GPT-3 to understand the answer candidates. Second, without image captions, Prophet still works steadily. This reflects the fact that our answer heuristics in prompts already provide sufficient information for Prophet to solve the task. Third, the prompt head is of less importance, indicating that GPT-3 is capable of understanding the task directly from the in-context examples. Finally, introducing extra information like object tags leads to a slight performance drop, which is contrary to the results in PICa. We conjecture this information has already been encoded in answer heuristics implicitly. 
\vspace{5pt}
\\
\noindent\textbf{Prediction behaviors in different stages.} In Table \ref{table:vqa_model}, we can observe a significant performance improvement of Prophet (stage-2) over its corresponding MCAN model (stage-1). 
To better understand this improvement, we conduct a statistical analysis of Prophet's prediction behaviors. As Prophet takes $K$ answer candidates from MCAN as inputs, we define three prediction behaviors of Prophet, namely ``keep top 1'', ``in top 2-$K$'', and ``beyond top $K$'', which means LLM’s prediction hits the top-1 candidate, one of the top 2-$K$ candidates, none of the $K$ candidates, respectively. Note that all the testing samples can be categorized into one of the three classes above. The statistical results in Fig.  \ref{fig:stat} show that: (i) for 68.1\% of the testing samples (green slice), Prophet keeps the top-1 predictions of MCAN. These samples achieve a 69\% accuracy and are mostly easy samples; (ii) for 21.8\% of the testing samples (blue slice), Prophet selects answers from the top 2-$K$ answer candidates. These samples are relatively hard, so that MCAN delivers a 24\% accuracy while Prophet has a much higher 40\% accuracy; (iii) for the remaining 10.1\% of the testing samples (yellow slice), Prophet resorts to the LLM to predict new answers beyond the answer candidates\footnote{The probability that Prophet's prediction is constituted of the combination of candidates is rare that can be neglected.}. For these challenging samples, MCAN only delivers a 12\% accuracy while Prophet magnificently achieves a 42\% accuracy. More analyses are provided in the appendix.
\begin{table}
	\centering
	\renewcommand\arraystretch{1.2}
	\begin{tabular}{lcc}
		LLM (version or size)~~~~~~~~~~~~~~~~~~~ &\makecell{per-sample\\average cost}& accuracy \\
		\ChangeRT{1.2pt}
		\multicolumn{1}{l}{\textit{commercial models}} \\
		GPT-3 (text-davinci-002) & \$0.2 & \multicolumn{1}{>{\columncolor{mygray}}c}{\textbf{60.8}} \\
		GPT-3 (3.5-turbo-instruct) & \$0.015 & 58.9\\
		\ChangeRT{0.5pt}
		\multicolumn{1}{l}{\textit{open-source models}} \\
		LLaMA-2 (7B) \cite{llama2}  & 2.7s & 56.6\\
		LLaMA-2-Chat (7B) \cite{llama2}&2.7s &54.0 \\
		LLaMA-2 (13B) \cite{llama2} & 4.8s & 57.9\\
		LLaMA-2-Chat (13B) \cite{llama2} & 4.8 & 56.5\\
		Mistral (7B) \cite{mistral}  &3.0s & 59.7\\ 
	\end{tabular}
	\caption{\textbf{Ablation study of different LLMs.} All variants use the default settings and are evaluated on the testing set of OK-VQA. The per-sample average costs of the open-source models are measured by the GPU running time on a server with A100 GPUs while the costs of the commercial models are measured by money.}\label{tab:aba_llms}
\end{table}
\begin{table}
	\centering
	\renewcommand\arraystretch{1.2}
	\begin{tabular}{l|cc|cc|c}
		\multirow{3}{*}{method} & \multicolumn{2}{c|}{stage-1}& \multicolumn{2}{c|}{stage-2} & \multirow{3}{*}{accu.} \\
		&\makecell{VQA\\model} & \makecell{LMM} & \makecell{LLM/\\LMM} & \makecell{pure text/\\img+text} & \\
		\ChangeRT{1.2pt}
		GPT-4o &-& - & - & - & 46.0\\
		Prophet& MCAN & -& GPT-4o & pure-text & 62.9\\
		\hline
		\multirow{2}{*}{Prophet++} & MCAN & GPT-4o &GPT-4o& pure-text & 63.8\\
		& MCAN & GPT-4o &GPT-4o& img+text & \textbf{64.2} \\
	\end{tabular}
	\caption{\textbf{Ablation study for Prophet++} on the testing set of OK-VQA. The first split contains the results of direct-prediction and Prophet-based prediction using the same LMM (GPT-4o). The second split contains two Prophet++ variants with different settings. $N$=16 is used for Prophet and Prophet++.}\label{tab:aba_lmms}
	\vspace{-5pt}
\end{table}
\vspace{5pt}
\\
\noindent\textbf{Different LLMs.} In Table \ref{tab:aba_llms}, we investigate the effects of different LLMs by replacing the default GPT-3 (text-davinci-002) with the latest commercial and open-source models. From the results, we have the following observations: (i) the capability of the default GPT-3 model consistently outperforms all the compared LLMs, including its accelerated variant (3.5-turbo-instruct); (ii) for the LLMs of the same version but different sizes (\emph{e.g.}, 7B and 13B LLaMA-2 models \cite{llama2}), the large-size ones show better performance than the small-size ones at the expense of near-linearly increasing running time; (iii) the chat-oriented variants like LLaMA-2-Chat \cite{llama2}, which are additionally trained by instruction tuning and human feedback \cite{instructgpt}, deliver inferior performance to their non-chatty counterparts. This can be explained by the introduced \emph{alignment tax} when aligning the model with human behaviors by RLHF; (iv) with only 7B model size, the latest LLM Mistral \cite{mistral} reports near \mbox{GPT-3} level performance, revealing the potential of open-source LLMs.
\vspace{5pt}
\\
\noindent\textbf{Effect of LMMs in Prophet++.} In Table \ref{tab:aba_lmms}, we investigate the effects of different strategies. The results suggest that: (i) although GPT-4o is a highly-capable LMM, directly using it in a direct-prediction manner achieves much worse performance than the Prophet-based counterpart (46.0 \emph{vs}. 62.9), showing the effectiveness of Prophet's two-stage prompting paradigm; (ii) by introducing the third type of answer heuristic in the first stage, Prophet++ delivers a 0.9-point improvement over Prophet, showing the effectiveness and complementarity of the answer-aware rationales; (iii) when visual signals are additionally utilized in the second stage, we obtain further a 0.4-point improvement, validating the effectiveness of multimodal prompting in Prophet++. More results with the latest LMMs are provided in the appendix.

\subsection{Main Results}\label{sec:main_result}
For the comparisons below, we use all the default settings except the number of examples $N$. We set $N$=20 for OK-VQA and A-OKVQA and respectively set $N$=7 and $N$=16 for ScienceQA and TextVQA as they need extra hint and OCR tokens. By instantiating Prophet with two VQA models, we obtain Prophet (MCAN) and Prophet (mPLUG). Finally, we report the result of Prophet++ (mPLUG) on all the datasets to compare with Prophet (mPLUG). 
\vspace{5pt}
\\
\noindent\textbf{Comparative results on OK-VQA.} Table \ref{sota} contains the comparisons of our Prophet and existing state-of-the-art methods on OK-VQA. The table is split into three sections. The first section lists the retrieval-based methods leveraging external KBs \cite{krisp,mavex,trig}. The
second section contains the methods that are directly pretrained on a large-scale multimodal corpus \cite{unifiedio,flamingo,pali}. The last two sections show the methods that incorporate the GPT-3 and GPT-4o \cite{pica,kat,revive,promptcap}, which are accessible via online APIs. 
\begin{table}
	\centering
	\renewcommand\arraystretch{1.2}
	\begin{tabular}{lc}
		method & accuracy \\
		\ChangeRT{1.2pt}
		\multicolumn{1}{l}{\textit{methods with external knowledge bases~~~~~~~~~~~~~~~~~~~~~~~~~}} \\
		KRISP \cite{krisp} & 38.9 \\ 
		Visual Retriever-Reader \cite{vrr}  & 39.2 \\ 
		MAVEx \cite{mavex}& 40.3 \\ 
		UnifER \cite{unifer} & 42.1 \\
		TRiG \cite{trig}  & 49.4 \\
		\ChangeRT{0.5pt}
		\multicolumn{1}{l}{\textit{methods with multimodal pretraining}} \\
		Unified-IO (2.8B) \cite{unifiedio}  & 54.0 \\ 
		Flamingo (80B) \cite{flamingo}  & 57.8 \\
		PALI (17B) \cite{pali} & {64.5}\\
		\ChangeRT{0.5pt}
		\multicolumn{1}{l}{\textit{methods with GPT-3 API}} \\
		PICa \cite{pica} & 48.0 \\ 
		KAT$^\dag$ \cite{kat} & 53.1 \\ 
		REVIVE$^\dag$ \cite{revive}  & 56.6 \\
		PromptCap (OFA)$^\dag$ \cite{promptcap}  & 60.4 \\
		\textbf{Prophet (MCAN)} & {61.1} \\
		\textbf{Prophet (mPLUG)} & {62.5} \\
		\ChangeRT{0.5pt}
		\multicolumn{1}{l}{\textit{methods with GPT-4o API}} \\
		\textbf{Prophet++ (mPLUG)} & \textbf{65.7} \\ 
	\end{tabular}
	\caption{\textbf{Comparisons to the state-of-the-art methods on OK-VQA} testing set. The compared methods are split into three groups based on their knowledge resources and usages.
		$^\dag$: method needs to query GPT-3 during training.}\label{sota}
	\vspace{-5pt}
\end{table}

Our Prophet belongs to the last section. It outperforms all the compared methods by a distinct margin. Prophet is 13.1 points higher than PICa \cite{pica} when both methods use GPT-3 as the only knowledge resource. This confirms our hypothesis that the capacity of GPT-3 has not been fully activated in previous studies. Compared to KAT \cite{kat} and REVIVE \cite{revive}, which utilize GPT-3 and other external KBs together in sophisticated systems, our Prophet is much simpler and more effective. Moreover, KAT, REVIVE, and PromptCap need to use GPT-3 to process all the training samples for their model training, which significantly increases the costs. In contrast, our Prophet only uses GPT-3 at inference time, which is more economical. Compared to the Flamingo-80B equipped with 32 in-context examples \cite{flamingo}, Prophet (MCAN) delivers a significant performance improvement. Despite the fact that Prophet (MCAN) has a clear performance gap compared to PALI-17B \cite{pali}, Prophet is more resource-efficient from the perspective of {reproducibility}\footnote{Flamingo-80B is trained on 1,536 TPUv4 for 15 days and PALI is trained on 1,024 TPUv4 for 7 days, which are unaffordable for most researchers. In contrast, Prophet (MCAN) uses one RTX-3090 to train a VQA model for 4 days and a certain number of GPT-3 invocations.}. Moreover, by replacing MCAN with the pretrained generative model mPLUG, our method exhibits a 1.4-point further improvement, showing the substantial contribution of a powerful VQA model for Prophet. Finally, Prophet++ (mPLUG) equipped with a modern LMM delivers a 3.2-point improvement over Prophet (mPLUG).
\begin{table}
\centering
\renewcommand{\arraystretch}{1.2} 
\begin{tabular}{l|cc|cc}
	\multirow{2}{*}{method}     & \multicolumn{2}{c|}{DA} & \multicolumn{2}{c}{MC} \\
	& \makebox[0.048\textwidth][c]{val} & test & \makebox[0.048\textwidth][c]{val} & test \\
	\ChangeRT{1.2pt}
	ClipCap    \cite{a-okvqa}   & 30.9 & 25.9 & 56.9 & 51.4 \\
	ViLBERT    \cite{a-okvqa}   & 30.6 & 25.9 & 49.1 & 41.5 \\
	LXMERT     \cite{a-okvqa}   & 30.7 & 25.9 & 51.4 & 41.6 \\
	KRISP      \cite{a-okvqa}   & 33.7 & 27.1 & 51.9 & 42.2 \\
	GPV-2      \cite{a-okvqa}   & 48.6 & 40.7 & 60.3 & 53.7 \\
	Unified-IO \cite{unifiedio} \enspace\enspace\enspace\enspace\enspace\enspace & -    & 45.2 & -    & - \\
	MCAN \cite{mcan} & 52.0 & 45.6 & - & -\\
	mPLUG \cite{mplug} & 59.1& 55.7& - & -\\
	PromptCap (OFA) \cite{promptcap} ~~~~~~& 56.3 &{59.6}&73.2 &73.1\\
	\ChangeRT{0.5pt}
	\textbf{Prophet (MCAN)} & {58.2} & {55.7} & {76.4} & {73.6} \\
	\textbf{Prophet (mPLUG)} & {64.7} & {58.5} & {76.6} & {75.1}\\
	\textbf{Prophet++ (mPLUG)} & \textbf{68.3} & \textbf{68.0} & \textbf{87.7} & \textbf{86.7}\\
\end{tabular}
\caption{\textbf{Comparisons to the state-of-the-art methods on A-OKVQA}. DA and MC refer to the direct-answer and multiple-choice tasks, respectively. For the MC task, we devise a Prophet variant with a slightly different prompt.}
\label{aok}
\end{table}
\begin{table}
	\renewcommand\arraystretch{1.2}
	\begin{subtable}[t]{0.47\columnwidth}
		\centering
		\begin{tabular}{lc}
			method & accu. \\
			\ChangeRT{1.2pt}
			MCAN \cite{mcan} & 51.2 \\
			mPLUG \cite{mplug} & 77.0 \\
			InstructBLIP \cite{instructblip} & 79.5 \\
			LLaMA-Adapter \cite{llamaadapter} & 80.3 \\
			MM-CoT \cite{mmcot} & 82.9\\
			Human Average \cite{scienceqa} & 87.5\\
			LLaVa \cite{llava} & 88.0 \\
			\hline
			\textbf{Prophet (mPLUG)} & {88.2}\\
			\textbf{Prophet++ (mPLUG)} & \textbf{90.5}\\
		\end{tabular}
		\captionsetup{font=footnotesize}
		\subcaption{ScienceQA (IMG)}
		\label{table:sciqa}
	\end{subtable}
	\quad
	\renewcommand\arraystretch{1.2}
	\begin{subtable}[t]{0.47\columnwidth}
		\centering
		\begin{tabular}{lc}
			method & accu. \\
			\ChangeRT{1.2pt}
			LoRRA \cite{textvqa} & 27.6\\
			M4C \cite{m4c} & 40.5\\
			PromptCap \cite{promptcap} & 51.9 \\
			mPLUG \cite{mplug} & 53.5\\
			TAP \cite{tap}& 54.0\\
			Flamingo-80B \cite{flamingo} & 54.1 \\
			LaTr \cite{latr} & 59.6 \\
			\hline
			\textbf{Prophet (mPLUG)} & {61.3}\\
			\textbf{Prophet++ (mPLUG)} & \textbf{61.8}\\
		\end{tabular}
		\captionsetup{font=footnotesize}
		\subcaption{TextVQA }
		\label{table:textvqa}
	\end{subtable}
	\caption{\textbf{Comparisons to the state-of-the-art methods} on ScienceQA and TextVQA.}
	\label{table:more-results}
	\vspace{-5pt}
\end{table}
\vspace{5pt}
\\
\noindent\textbf{Comparative results on A-OKVQA.} Table \ref{aok} contains the comparative results on the challenging A-OKVQA dataset. The results on the DA task show that Prophet (MCAN) model significantly outperform the existing approaches including its base VQA model MCAN, showing its effectiveness and generalizability. Compared to the state-of-the-art method PromptCap \cite{promptcap} which also involves a pretrained VQA model OFA \cite{ofa} and \mbox{GPT-3}, Prophet (MCAN) exhibits similar performance when using a weaker VQA model.  

For the MC task, we introduce a Prophet variant with slightly modifying the prompt used in the original Prophet. In particular, we add the multiple-choice information into both the in-context examples and testing input to instruct GPT-3 to \emph{choose} the correct one from four choices.
Compared with all the methods, Prophet (MCAN) surpasses all the counterparts on the MC task, showing the flexibility and scalability of Prophet. Moreover, by introducing a more powerful VQA model mPLUG, Prophet (mPLUG) consistently outperforms Prophet (MCAN), while Prophet++ (mPLUG) further outperforms Prophet (mPLUG). 
\vspace{5pt}
\\
\noindent\textbf{Results on ScienceQA and TextVQA}
To verify the generalization ability of Prophet, we conduct experiments on two additional knowledge-based VQA datasets ScienceQA (IMG) and TextVQA, which require different types of knowledge (\emph{i.e.}, scientific knowledge and OCR knowledge) than that for OK-VQA and A-OKVQA. 
A
Table \ref{table:more-results} shows that comparative results of Prophet and existing state-of-the-art methods on respective datasets. As we have witnessed the steady improvements of mPLUG over MCAN, we only report the results for Prophet (mPLUG) on these two datasets. Specifically, Prophet surpasses all the counterparts on ScienceQA (IMG), including the base VQA model mPLUG, the average human performance \cite{scienceqa}, and the latest LMM LLaVa trained with visual instruction tuning \cite{llava}. On TextVQA, Prophet outperforms the published state-of-the-art methods, including the methods with text-aware or layout-aware pretraining on large-scale scene-text image datasets \cite{latr,tap}. Finally, the Prophet++ models prominently outperform their corresponding Prophet counterparts on respective datasets.

\section{Conclusion}
In this paper, we present Prophet---a conceptually simple framework which uses LLMs as the knowledge engine for knowledge-based VQA. To better activate the few-shot learning capacity of LLMs, we introduce a novel paradigm to prompt LLMs with complementary answer heuristics. Extensive ablations, comparative experiments, and comprehensive analyses on four diverse knowledge-based VQA datasets show the superiority of Prophet over all existing state-of-the-art methods. Notably, Prophet can be integrated with varied combinations of VQA models, LLMs, and even LMMs, showing the flexibility, scalability, and generalizability of our framework. 

\appendices

\section{Broader Impact}
From the larger multimodal model (LMM) point of view, Prophet is a loosely-coupled LMM consisting of a vision-language (VL) model and a frozen LLM, aiming to endow the VL model with knowledge reasoning ability. Compared with the tightly-coupled LMMs (\emph{e.g.}, Flamingo \cite{flamingo} and LLaVa \cite{llava}) which jointly optimize the VL model and LLM in an end-to-end manner, Prophet is more flexible that can support any open-source or commercial LLM. 

Moreover, Prophet can also be regarded as a \emph{learning-to-prompt} paradigm that learns an external model to generate prompts to better comprehend the target task, thus facilitating the capability of the pretrained LLM (or LMM). From this point of view, recent studies like VoxPoser \cite{voxposer} and SoM-Prompting \cite{somprompt} share a similar idea with our work. We believe this paradigm can be widely used in a variety of LLM-related tasks. 

\section{More Implementation Details}
\subsection{The Default VQA Model}

Our default VQA model is carefully designed in terms of model architecture and training strategy. In the following table, we show the improvements of our default MCAN model over the counterparts trained from scratch. More details are provided next.
\begin{table}[H]
	\small
	\renewcommand\arraystretch{1.2}
	\centering
	\begin{tabular}{ccc}
		\makecell{from scratch,\\original model \cite{mcan}}  & \makecell{from scratch, \\improved model}  & \makecell{transfer learning,\\improved model} \\
		\ChangeRT{1.2pt}
		31.5 & 35.6 & \textbf{53.0} \\
	\end{tabular}
	\vspace{-10pt}
\end{table}

\noindent\textbf{Improved model architecture.} We introduce an improved variant of MCAN \cite{mcan} based on its open-sourced MCAN-large implementation. Our modifications to the model architecture include: (i) we replace the original bottom-up-attention features with the grid-based features extracted from the CLIP's visual encoder with RN50$\times$64 backbone \cite{clip}; (ii) we introduce the RoPE mechanism \cite{rope} to each image self-attention layer of MCAN to supplement the grid-based features with positional information; and (iii) we replace the original LSTM network with a pre-trained BERT-large model \cite{bert} as the text encoder before MCAN. Table \ref{tab:arch} shows the accuracies of different model variants on the testing set of OK-VQA. By progressively adding the modifications to the original MCAN model, our improved MCAN model reports a 53.0\% accuracy, which is on par with current state-of-the-art methods like KAT \cite{kat}.

\begin{table}[h]
	\centering
	\renewcommand{\arraystretch}{1.2} 
	\begin{tabular}{lc}
		case & OK-VQA accuracy        \\
		\ChangeRT{1.2pt}
		original MCAN            & 43.6   \\ 
		+ CLIP visual feats      & 49.6   \\
		+ RoPE mechanism         & 50.3   \\
		+ BERT as the text encoder  ~~~~~~~~~~~~~~~~~ & \textbf{53.0}   \\
	\end{tabular}
	\caption{\textbf{Ablations for model architectures.} `+' denotes each modification is applied to the previous variant.}
	\label{tab:arch}
\end{table}

\begin{table}
	\centering
	\renewcommand{\arraystretch}{1.2} 
	\begin{tabular}{l|cc}
		config & setting \\
		\ChangeRT{1.2pt}
		optimizer & AdamW \\
		weight decay & 0.01 \\
		optimizer momentum & $\beta_1, \beta_2{=}0.9, 0.98$ \\
		batch size & 64 \\
		warm-up learning rate & 1e-3 \\
		warm-up strategy & only update new parameters\\
		warm-up epochs & 1 \\
		base learning rate & 5e-5 \\
		learning rate schedule & step decay \\
		learning rate decay rate & 0.2 \\
		learning rate decay epoch~~~ & 6 \\
		total training epochs & 6 \\
	\end{tabular}
	\vspace{-5pt}
	\caption{\textbf{Training settings.} These hyper-parameters are used in our experiments.} \label{tab:finetune_setting} 
	\vspace{-5pt}
\end{table}

\begin{table}[H]
	\centering
	\renewcommand{\arraystretch}{1.2} 
	\begin{tabular}{lc}
		training strategy & OK-VQA accuracy \\
		\ChangeRT{1.2pt}
		(a) train from scratch             & 35.6 \\
		(b) pretrain, w/o finetune     & 41.1 \\ 
		(c) w/ finetune, replace last layer  & 47.7 \\ 
		(d) w/ finetune, append new answers  & \textbf{53.0}  \\
	\end{tabular}
	\caption{\textbf{Ablations for training strategies.} All variants use the improved model architecture in the last row in Table \ref{tab:arch}.
	}    
	\label{tab:train_strategy}
\end{table}

\begin{table*}
	\centering
	\begin{tabular}{|l|}
		\hline
		\texttt{Please answer the question according to the context and the answer candidates. Each answer candidate is associated with a ~~~}\\
		\texttt{confidence score within a bracket. The true answer may not be included in the candidates.}
		\\
		\texttt{===}\\
		\texttt{Context: The motorcycle racers are getting ready for a race.}\\
		\texttt{===}\\
		\texttt{Question: What sport are these guys doing?}\\
		\texttt{===}\\
		\texttt{Candidates: motorcross(0.94), motocross(0.79), bike(0.35), dirt bike(0.28), motorcycle(0.03),}\\
		\texttt{bmx(0.03), cycling(0.02), motorbike(0.02), race(0.02), bicycle(0.02)}\\
		\texttt{===}\\
		\texttt{Answer: motorcross}
		\\
		\texttt{===}\\
		\texttt{Context: a black motorcycle parked in a parking lot.}\\
		\texttt{===}\\
		\texttt{Question: What sport can you use this for?}\\
		\texttt{===}\\
		\texttt{Candidates: race(0.53), motorcycle(0.41), motocross(0.19), bike(0.17), motorcross(0.15),}\\
		\texttt{cycling(0.11), dirt bike(0.10), ride(0.08), bicycling(0.01), bicycle(0.01)}\\
		\texttt{===}\\
		\texttt{Answer:}\\
		\hline
	\end{tabular}
	\caption{\textbf{An exemplar prompt for the standard Prophet.} We show \emph{one} in-context example here due to space limitations. Following the implementations in PICa \cite{pica} and KAT \cite{kat}, we use a special symbol `\texttt{===}' to separate each two lines.}
	\label{tab:std_prompt}
\end{table*}

\begin{table*}
	\centering
	\begin{tabular}{|l|}
		\hline
		\texttt{\textcolor{red}{Please choose the correct answer in the choices} according to the context, the question and the answer candidates. Each answer}\\
		\texttt{candidate is associated with a confidence score within a bracket. The true answer may not be included in the candidates.}
		\\
		\texttt{===}\\
		\texttt{Context: A young man riding a skateboard on a sidewalk.}\\
		\texttt{===}\\
		\texttt{Question: What part of his body will be most harmed by the item in his mouth?}\\
		\texttt{===}\\
		\texttt{Candidates: skateboard(0.02), nothing(0.02), table(0.01), leg(0.01), helmet(0.00), knees(0.00),}\\
		\texttt{skateboarding(0.00), head(0.00), teeth(0.00), falling(0.00)}\\
		\texttt{===}\\
		\texttt{\textcolor{red}{Choices: (A) back, (B) lungs, (C) feet, (D) eyes}}\\
		\texttt{===}\\
		\texttt{Answer: \textcolor{red}{(B)}}
		\\
		\texttt{===}\\
		\texttt{Context: a young boy kneeling on a skateboard on the street.}\\
		\texttt{===}\\
		\texttt{Question: What did this lad likely injure here?}\\
		\texttt{===}\\
		\texttt{Candidates: skateboard(0.18), shoes(0.02), shoe(0.02), skateboarding(0.01), street(0.01),}\\
		\texttt{flowers(0.01), skating(0.01), boy(0.01), head(0.00), skateboarder(0.00)}\\
		\texttt{===}\\
		\texttt{Choices: \textcolor{red}{(A) knee, (B) elbow, (C) rear, (D) board}}\\
		\texttt{===}\\
		\texttt{Answer:}\\
		\hline
	\end{tabular}
	\caption{\textbf{An exemplar prompt for the Prophet variant on the MC task of A-OKVQA.} Compared to the standard prompt in Table \ref{tab:std_prompt}, we add one extra line of choices for the example and testing input, and change the output format to adapt to the multiple-choice task. All the differences are marked in \textcolor{red}{red}.
	}
	\label{tab:aokmc_prompt}
\end{table*}

\begin{table*}
	\centering
	\begin{tabular}{|l|}
		\hline
		\texttt{Please choose the correct answer in the choices according to the context, the question and the answer candidates. Each answer}\\ 
		\texttt{candidate is associated with a confidence score within a bracket. The true answer may not be included in the candidates.}\\
		\texttt{===}\\
		\texttt{Context: A picture of a black and white model of a molecule. \textcolor{red}{The model below represents graphite. Graphite is used to make}}\\ \texttt{\textcolor{red}{pencil lead.}}\\
		\texttt{===}\\
		\texttt{Question: Complete the statement. Graphite is ().}\\
		\texttt{===}\\
		\texttt{Candidates: an elementary substance(1.00), a compound(0.02), an adult substance(0.01), an an elementary substance(0.01)}\\
		\texttt{===}\\
		\texttt{Choices: (A) a compound, (B) an elementary substance}\\
		\texttt{===}\\
		\texttt{Answer: (B)}
		\\
		\texttt{===}\\
		\texttt{Context: A pair of eye glasses with the word h on them. \textcolor{red}{The model below represents a molecule of hydrogen. Hydrogen gas was}}\\ \texttt{\textcolor{red}{once used to make large airships, such as blimps, float. It is no longer used in airships because it catches fire easily.}}\\
		\texttt{===}\\
		\texttt{Question:Complete the statement. Hydrogen is ().}\\
		\texttt{===}\\
		\texttt{Candidates: a compound(0.68), an elementary substance(0.32), the same substance(0.00), the same amount(0.00)}\\
		\texttt{===}\\
		\texttt{Choices: (A) an elementary substance, (B) a compound}\\
		\texttt{===}\\
		\texttt{Answer:}\\
		\hline
	\end{tabular}
	\caption{\textbf{An exemplar prompt for the Prophet variant on ScienceQA (IMG).} The sentences marked in \textcolor{red}{red} are the optional text hints provided by the dataset.}
	\label{tab:sciqa_prompt}
\end{table*}

\begin{table*}
	\centering
	\begin{tabular}{|l|}
		\hline
		\texttt{Please answer the question according to the context and the answer candidates. Each answer candidate is associated with a ~~~}\\
		\texttt{confidence score within a bracket. The true answer may not be included in the candidates.}
		\\
		\texttt{===}\\
		\texttt{Context: A close up of a cell phone with a keyboard.}\\
		\texttt{===}\\
		\texttt{\textcolor{red}{OCR: Market, 3, Facebook, Browser, 5, 4, 6, 1, 8, 30.}}\\
		\texttt{===}\\
		\texttt{Question: How many apps are on this page excluding market?}\\
		\texttt{===}\\
		\texttt{Candidates: 6(0.20), 5(0.19), 8(0.18), 9(0.12), 7(0.08), answering does(0.05),10(0.05),13(0.05),12(0.04),4(0.04)}\\
		\texttt{===}\\
		\texttt{Answer: 7}
		\\
		\texttt{===}\\
		\texttt{Context: A screenshot of a yahoo mail page.}\\
		\texttt{===}\\
		\texttt{\textcolor{red}{OCR: Free, Page, Nake WT My Page, ADVERTISEMENT, YAHOO!, FREE Camera Phone, Notepad, MAIL, Yaboo! Mail.}}\\
		\texttt{===}\\
		\texttt{Question: What is free on this page?}\\
		\texttt{===}\\
		\texttt{Candidates: amera(0.40), video camera(0.29), video(0.13), photos(0.04), video call(0.04), webcam(0.03), videos(0.03),}\\ \texttt{photography(0.01), photoshop(0.01), internet explorer(0.01)}\\
		\texttt{===}\\
		\texttt{Answer:}\\
		\hline
	\end{tabular}
	\caption{\textbf{An exemplar prompt for the Prophet variant on TextVQA.} Compared to the standard prompt, we additionally introduce the OCR tokens (marked in \textcolor{red}{red}) extracted from an off-the-shelf OCR system.  }
	\label{tab:textvqa_prompt}
\end{table*}

\noindent\textbf{Training recipe.} We first pretrain the model on the augmented \emph{train+val+vg} dataset from VQAv2 \cite{vqa2} and Visual Genome\cite{visualgenome}, with excluding the samples whose images are used in the testing split of OK-VQA to avoid data contamination. The settings for the pretraining stage are identical to the original implementation of MCAN. 
After that, the model is finetuned on the downstream OK-VQA and A-OKVQA datasets, respectively. For finetuning, the commonly used strategy is to replace the last linear layer (\emph{i.e.}, the classification layer) with a new layer to adapt to the answer vocabulary of the downstream dataset. However, the answer vocabularies of the pretraining and finetuning datasets are \emph{partially} overlapped. To maximally utilize the pretrained model parameters in the last layer, we inherit the parameters of existing answers and append new parameters for the new answers. After that, we freeze all the pretrained parameters and only update the new parameters for one epoch as a warm-up, and then train all model parameters for the rest training epochs. 
The detailed settings for the finetuning stage are shown in Table \ref{tab:finetune_setting}.

Table \ref{tab:train_strategy} shows the effects of different training strategies. Even without finetuning, the pretrained model (b) is superior to the model trained from scratch (a), implying the importance of pretraining. Moreover, our new finetuning strategy (d) leads to significantly better performance than the commonly used strategy (c), showing the effectiveness of inheriting model parameters for existing answers.

\subsection{Prompt Formats}
We show an exemplar prompt for the standard Prophet in Table \ref{tab:std_prompt} and an exemplar prompt for the variant designed for the MC task of A-OKVQA in Table \ref{tab:aokmc_prompt}. 
The exemplar prompts for ScienceQA and TextVQA are illustrated in Table \ref{tab:sciqa_prompt} and \ref{tab:textvqa_prompt}, respectively.

\section{More Quantitative and Qualitative Analyses}
We provide more in-depth analyses of Prophet's performance on the testing set of OKVQA. All results are carried out using the default settings.

\begin{table}
	\centering
	\renewcommand{\arraystretch}{1.2} 
	\begin{tabular}{l|cc}
		category & MCAN & Prophet \\
		\ChangeRT{1.2pt}
		Plants and Animals & 52.58 & \textbf{63.67} \\
		Science and Technology & 48.10 & \textbf{48.81} \\
		Sports and Recreation & 59.08 & \textbf{66.00} \\
		Geography, History, Language and Culture & 52.48 & \textbf{62.98} \\
		Brands, Companies and Products & 51.98 & \textbf{54.77} \\
		Vehicles and Transportation & 50.82 & \textbf{58.01} \\
		Cooking and Food & 55.53 & \textbf{62.09} \\
		Weather and Climate & 65.12 & \textbf{68.37} \\
		People and Everyday life & 49.44 & \textbf{54.67} \\
		Objects, Material and Clothing & 50.05 & \textbf{57.20} \\
	\end{tabular}
	\caption{\textbf{Per-category accuracies } of MCAN (stage-1) and Prophet (stage-2). This performance improvements of using GPT-3 are observed on all categories. 
	} \label{tab:acc_per_category} 
\end{table}

\begin{table}
	\setlength{\arrayrulewidth}{1.2pt}
	\renewcommand{\arraystretch}{1.2} 
	\centering
	\begin{tabular}{c|cc}
		\diagbox{Stage 1 pred.}{Stage 2 pred.} & ~~~~correct~~~~ & ~~~~wrong~~~~ \\
		\ChangeRT{1.2pt}
		correct & 54.4\% & 4.2\% \\
		wrong & 12.0\% & 29.4\% \\
	\end{tabular}
	\caption{\textbf{Prophet's combinatorial prediction behaviors} in two stages. Prophet maintains the majority of correct predictions at stage-1, and the accuracy improvement by stage-2 is mainly because the number of \emph{wrong-to-correct} samples is larger than that of the \emph{correct-to-wrong} samples.
	}\label{tab:pred_before_after}
	\vspace{-5pt}
\end{table}

\subsection{Quantitative Analysis}
First, we show the per-type accuracies of MCAN (stage-1) and Prophet (stage-2) in Table \ref{tab:acc_per_category}. Prophet outperforms MCAN on all categories, indicating that generality of the knowledge in GPT-3. The improvement on the ``Science and Technology'' category is not as large as the rest categories. which can be explained that the required knowledge for this category is more specialized and professional. These questions are also challenging for humans.

Second, we calculate the distribution of four situations of the predictions from stage-1 and stage-2 in Table \ref{tab:pred_before_after}. From the results, we can see that: (i) Prophet maintains the majority of correct predictions by MCAN and only 4.2\% samples are overturned; (ii) the improvement of Prophet is mainly due to the fact that the proportion of \emph{wrong-to-correct} samples (12.4\%) is larger than that of the \emph{correct-to-wrong} samples (4.2\%); (iii) there are still a non-negligible amount of samples (29.4\%) that both MCAN and Prophet fail, which leaves sufficient room for future improvement.

\begin{table}
	\centering
	\renewcommand{\arraystretch}{1.2} 
	\begin{tabular}{l|c}
		failure cause & proportion \\
		\ChangeRT{1.2pt}
		(a) insufficient visual understanding & 27.3\% \\
		(b) incorrect knowledge reasoning & \textbf{44.1}\% \\
		(c) correct but differently expressed answer & 22.8\% \\
		(d) others & 5.8\% \\
	\end{tabular}
	\caption{\textbf{The distribution of failure causes} by human studies. 
	} \label{tab:cause_of_failure} 
	\vspace{-5pt}
\end{table}

Third, we perform human studies to analyze the causes of wrong predictions in Table \ref{tab:cause_of_failure}. For each category, we randomly sample 10\% testing samples that Prophet fails to get the correct answer. This results in 172 samples. We ask three annotators to categorize each sample into one of the following four failure causes: (a) insufficient visual understanding; (b) incorrect knowledge reasoning; (c) correct but differently expressed answer; (d) others (\emph{e.g.}, the failure is caused by the ambiguity of the question). From the results, we can see that the cause of ``(b) incorrect knowledge reasoning'' accounts for the highest proportion, which suggests that the bottleneck of Prophet still lies in the knowledge acquisition and reasoning. The cause of ``(a) insufficient visual understanding'' has the second highest proportion, showing the potential of devising more powerful VQA models. The cause of ``(c) correct but differently expressed answer'' also accounts for a considerable ratio. This reflects the limitation of the annotations and evaluation metric of OK-VQA.

\begin{table}
	\centering
	\renewcommand\arraystretch{1.2}
	\begin{tabular}{l|cc|cc|c}
		\multirow{3}{*}{method} & \multicolumn{2}{c|}{stage-1}& \multicolumn{2}{c|}{stage-2} & \multirow{3}{*}{accu.} \\
		&\makecell{VQA\\model} & \makecell{LMM} & \makecell{LLM/\\LMM} & \makecell{pure text/\\img+text} & \\
		\ChangeRT{1.2pt}
		GPT-4o &-& - & - & - & 46.0\\		
		L1.5-7B \cite{llava-1.5} &-& - & - & - & 58.3\\		
		Q2.5-7B \cite{qwen2.5vl} &-& - & - & - & 59.7\\
		\hline
		\multirow{3}{*}{Prophet}& MCAN & -& GPT-4o & pure-text & 62.9\\	
		& L1.5-7B & -& GPT-4o& pure-text & 66.5\\
		& Q2.5-7B & -& GPT-4o& pure-text & 66.4\\
		\hline
		\multirow{3}{*}{Prophet++} & MCAN & GPT-4o &GPT-4o& img+text & {64.2} \\
		& L1.5-7B & GPT-4o &GPT-4o& img+text & \textbf{68.3} \\
		& Q2.5-7B & GPT-4o &GPT-4o& img+text & {67.2} \\
	\end{tabular}
	\caption{\textbf{More ablation study of the latest LMMs in Prophet and Prophet++}. The first split contains the direct-prediction results of three state-of-the-art LMMs, namely GPT-4o, LLaVA-1.5-7B (L1.5-7B) \cite{llava-1.5}, and Qwen2.5-VL-7B (Q2.5-7B)\cite{qwen2.5vl}. The second and last splits contain different combinations of LMMs for Prophet and Prophet++, respectively. $N$=16 is used for Prophet and Prophet++.}\label{tab:more_aba_lmms}
	\vspace{-10pt}
\end{table}

Finally, we investigate the combinations of the latest LMMs for Prophet and Prophet++ in Table \ref{tab:more_aba_lmms}, which complements the results of Table 3 in the main text. The following observations are obtained from the results: 
(i) the direct-prediction performance by LLaVA-1.5-7B (L1.5-7B) \cite{llava-1.5} and Qwen2.5-VL-7B (Q2.5-7B) \cite{qwen2.5vl} significantly exceed that of GPT-4o, which may be explained by the fact that OK-VQA has been included in their training data; 
(ii) by utilizing the strong LMM L1.5-7B (or Q2.5-7B) as the VQA model, the resulting Prophet and Prophet++ significantly outperform their direct-prediction counterparts and MCAN-based counterparts, respectively, which verifies the effectiveness and generalizability of our frameworks to adapt to the latest LMMs;
(iii) although Q2.5-7B achieves slightly better direct-prediction accuracy than L1.5-7B (59.7 \emph{vs}. 58.3), its corresponding Prophet and Prophet++ models are inferior to the counterparts with L1.5-7B. This could be explained by the observation that the answer candidates generated by Q2.5-7B are more likely to be \emph{synonyms} with similar confidence scores, which may mislead the LLM/LMM in stage-2 and limit the final performance.

\begin{figure*}
	\begin{center}
		\includegraphics[width=\linewidth]{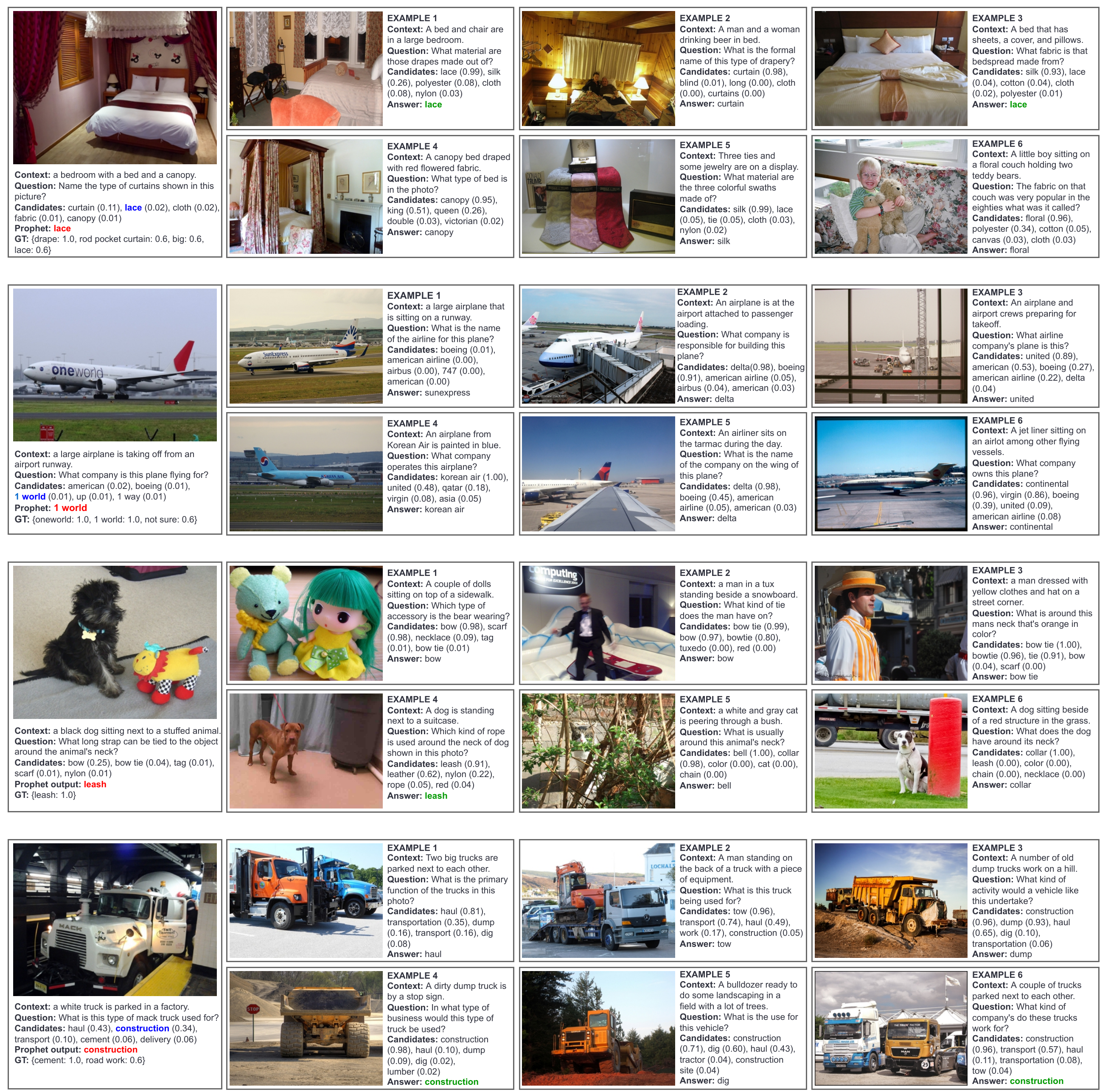}
		\caption{We show two typical samples consisting of the testing inputs (left) and their in-context examples (right). The \textcolor{red}{\textbf{predicted answers}} of Prophet have a high probability to appear in the \textcolor{blue}{\textbf{answer candidates}} and \textcolor[rgb]{0, 0.7, 0.31}{\textbf{answer-aware examples}}, showing the effectiveness of answer heuristics in enhancing LLM’s ability to predict the correct answer.}
		\label{fig:vis1}
		\vspace{-10pt}
	\end{center}
	\vspace{-5pt}
\end{figure*}

\begin{figure*}
	\begin{center}
		\includegraphics[width=0.95\linewidth]{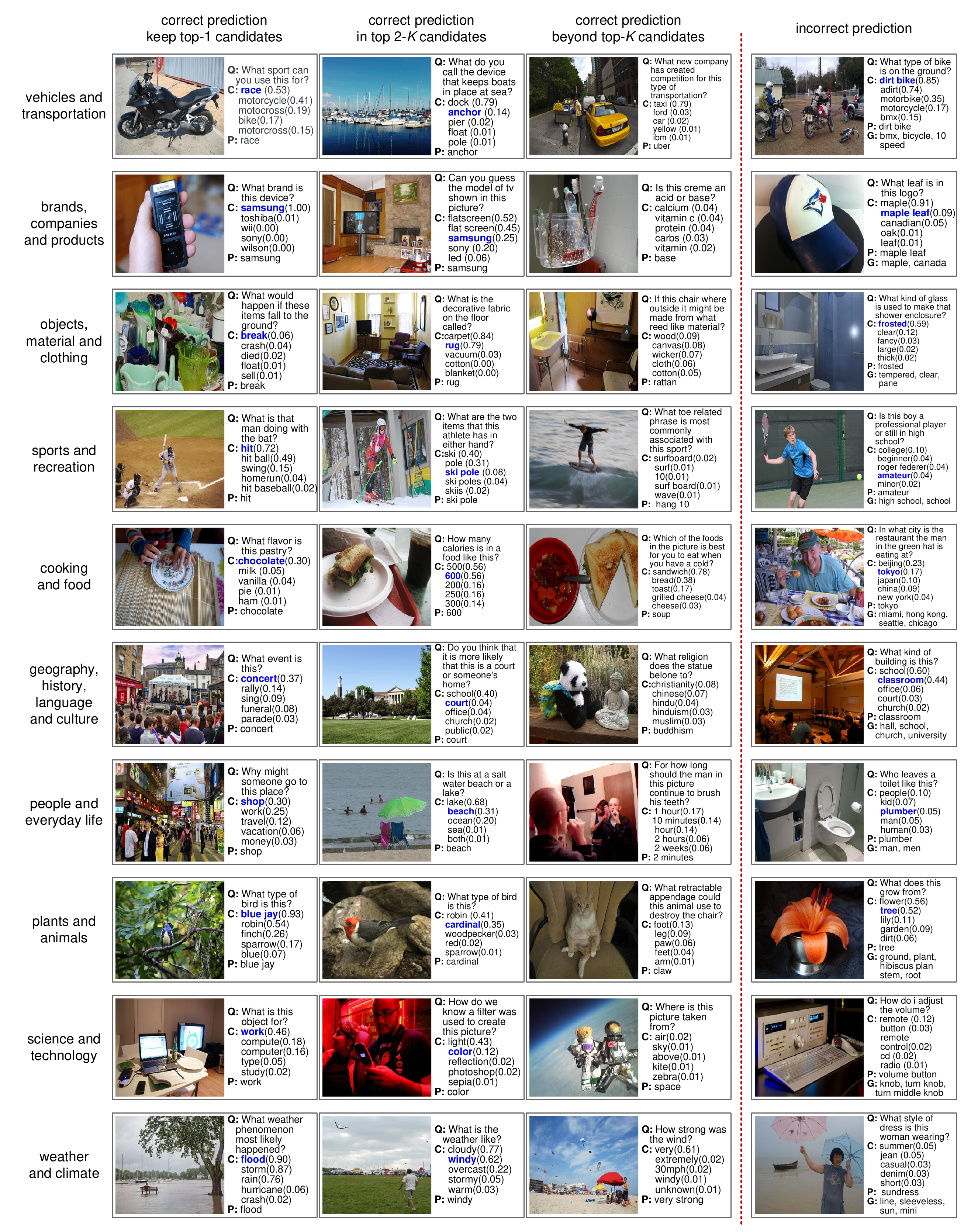}
		\caption{\textbf{Different categories and prediction behaviors.} 
			Each row contains four testing samples from a specific knowledge category. The first to the third columns correspond to the correctly answered samples of different prediction behaviors (\emph{i.e.}, keep top-1, in top 2-$K$, and beyond top-$K$). The last column contains failure samples.
		}\label{fig:per-cate-examples}
		\vspace{-15pt}
	\end{center}
\end{figure*}

\subsection{Qualitative Analysis}

In Fig. \ref{fig:vis1}, we illustrate two typical samples consisting of the testing inputs and their in-context examples to explain how the answer heuristics work. The results show that the synergy of answer candidates and the answer-aware examples facilitates the generation of high-quality answers. In the first sample, the candidate answer `\emph{lace}' with a low confidence score is finally selected by the LLM as it frequently appears in the in-context examples. In the second sample, we see that Prophet can make a correct prediction beyond the answer candidates when the proper answer heuristic (the word `\emph{leash}') is provided in the in-context examples.     

Fig. \ref{fig:per-cate-examples} demonstrates some testing samples from different knowledge categories. In the 1st-3rd columns, we show the correctly answered samples with different prediction behaviors (\emph{i.e.}, keep top-1, in top 2-$K$, and beyond top-$K$). The visualized results indicate that Prophet can adaptively choose suitable answers from candidates. In the last column, we show some failure samples, implying that there is still room for future improvement.

\bibliographystyle{IEEEtran}
\bibliography{prophet_oy}

\ifCLASSOPTIONcaptionsoff
  \newpage
\fi

\end{document}